\documentclass[runningheads]{llncs}

\usepackage{booktabs}
\usepackage{multirow}
\usepackage{siunitx}
\usepackage{eccv}

\usepackage{eccvabbrv}

\usepackage{graphicx}
\usepackage{booktabs}
\usepackage{colortbl}
\usepackage{xcolor}

\usepackage[accsupp]{axessibility}  

\usepackage[pagebackref,breaklinks,colorlinks,citecolor=eccvblue]{hyperref}
\usepackage{hyperref}
\usepackage{graphicx}
\usepackage{subcaption}
\usepackage{float} 
\usepackage{orcidlink}

\usepackage{xcolor}

\usepackage{algorithm}
\usepackage{algpseudocode}
\begin{document}

\title{Visual Semantic Entropy:\\ Do Vision Language Models Recognize Visual Ambiguity?}

\titlerunning{Visual Semantic Entropy}

\author{
Ta Duc Huy\inst{1}\orcidlink{0000-0001-6181-0478} \and
Trang Nguyen\inst{1}\\
Townim Chowdhury\inst{1}\orcidlink{0000-0003-1780-6046} \and
Ankit Yadav\inst{1}\orcidlink{0009-0006-8323-3919} \and
Minh-Son To\inst{2}\orcidlink{0000-0002-8060-6218} \\
Zhibin Liao\inst{1}\orcidlink{0000-0001-9965-4511} \and
Johan W.Verjans\inst{1}\orcidlink{0000-0002-8336-6774} \and
Vu Minh Hieu Phan\inst{1}\orcidlink{0000-0003-3861-0296}
}

\authorrunning{Ta Duc Huy et al.}

\institute{
Australian Institute for Machine Learning, Adelaide University, Australia \and
Flinders University, Australia\\
\email{tdh512194@gmail.com}
}

\maketitle

\begin{abstract}

Vision-language models can produce confident answers on visually ambiguous inputs, resulting in biased predictions. Common entropy-based methods, such as Semantic Entropy (SE), rely on output diversity. Yet our analysis shows that \emph{overconfident visual embeddings suppress output diversity} under stochastic decoding, causing SE to underestimate uncertainty in such cases. Recent methods instead probe output diversity through input perturbations, including textual paraphrasing or joint text-image perturbations, and show improved performance.
We study these approaches and reveals that the resulting variability is often dominated by textual changes rather than visual evidence, causing uncertainty estimates to reflect \emph{prompt sensitivity} rather than visual ambiguity.
We therefore propose \textbf{Visual Semantic Entropy (VSE)}, which perturbs \emph{only the image} to probe nearby visual variations while keeping the text query fixed. VSE measures uncertainty by clustering generated answers into semantic prototypes and computing the mass-weighted dispersion among them. Extensive evaluation across five modern vision-language models and five diverse VQA benchmarks demonstrates that VSE effectively captures visual ambiguity, establishing a new state-of-the-art for VLM uncertainty estimation. Code is available:
\href{https://github.com/tadeephuy/visual-semantic-entropy}{https://github.com/tadeephuy/visual-semantic-entropy}

\end{abstract}

\section{Introduction}

\begin{figure}[!]
    \centering
    \includegraphics[width=0.94\linewidth]{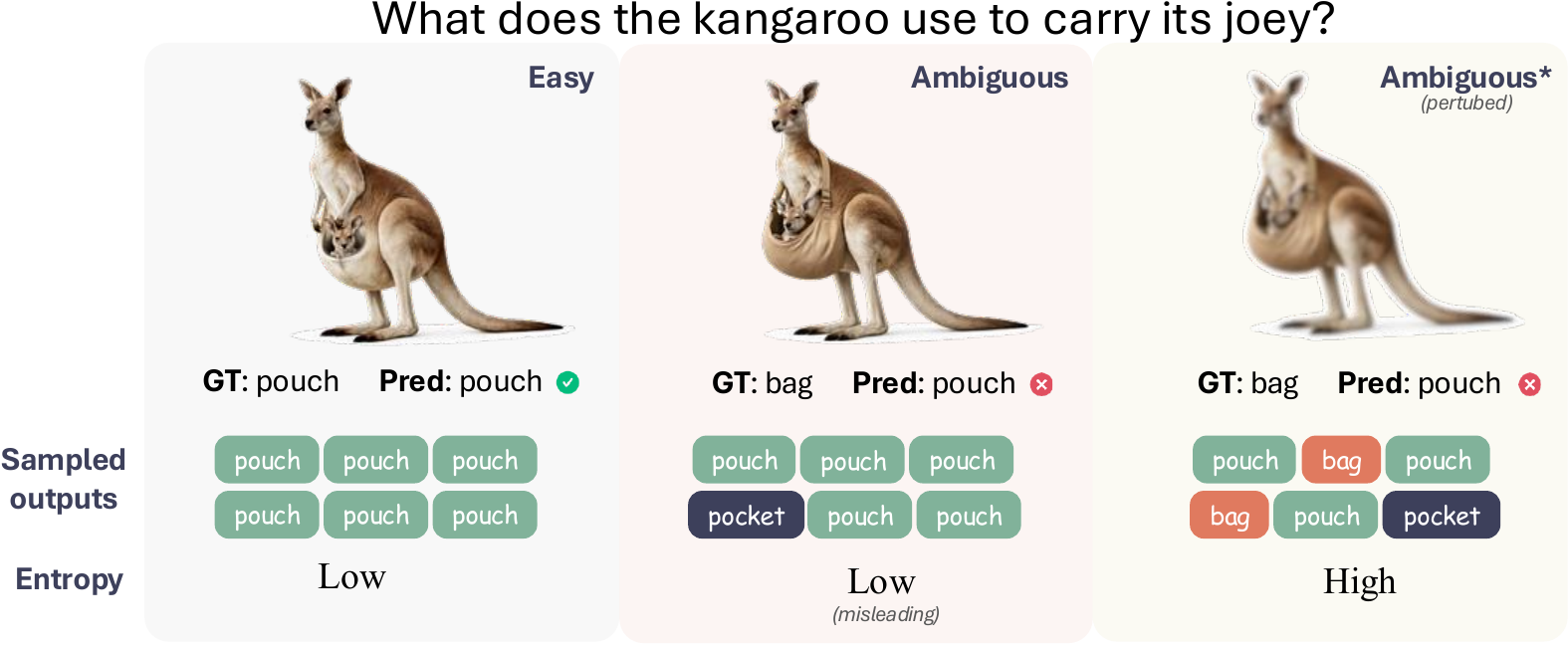}
    \caption{
\textbf{VLMs can be confidently wrong on ambiguous images.}
Sampling-based methods such as Semantic Entropy estimate uncertainty by repeatedly sampling model outputs and computing the entropy of the resulting answer distribution, where low entropy indicates high certainty.
\textit{Left (Easy):} For a visually clear image, the model correctly predicts ``pouch'' and repeated sampling consistently outputs ``pouch'', producing low entropy that appropriately reflects high certainty.
\textit{Middle (Ambiguous):} When the image is visually biased, the ground truth is ``bag'' but the model predicts ``pouch''. Despite this ambiguity, repeated sampling still yields only ``pouch'' resulting in low entropy, which is \textbf{misleading}.
\textit{Right (Ambiguous, perturbed):} After perturbing the image to probe alternative local views, sampling produces diverse answers such as ``pouch'', ``bag'', and ``pocket'', increasing entropy and exposing visual uncertainty.
}
    \label{fig:teaser}
\end{figure}

Vision-language models (VLMs) excel at visual question answering (VQA), a setting in which uncertainty can arise from ambiguity in the visual evidence. In this work, we study uncertainty estimation for VLMs in VQA. Reliable uncertainty estimates are essential for detecting unreliable predictions and enabling trustworthy deployment of VLMs. Current approaches to uncertainty estimation (UE) in VLMs include \emph{verbalized} methods that prompt models to report confidence~\cite{groot2024overconfidence,pelucchi2023exploring,valdenegro2021find,xuan2025seeing}, \emph{logit-based} methods derived from output probabilities or token entropy~\cite{li2024reference,huangrevisiting}, and \emph{consistency-based} methods that measure agreement across sampled generations~\cite{manakul2023selfcheckgpt,khan2024consistency,zhang2024radflag}. Another widely used class is \emph{entropy-based} methods, which estimate uncertainty from the semantic distribution of sampled answers, such as Semantic Entropy (SE) and its variants~\cite{farquhar2024detecting,nguyen2025beyond,nikitin2024kernel,zhang2024vl}.

In this paper, we set out to analyze existing UE methods and identify 3 limitations that motivate our approach:
\textbf{(1)} Overconfident visual representations can produce highly consistent outputs under sampling, causing entropy-based methods such as SE~\cite{farquhar2024detecting} to underestimate uncertainty.
\textbf{(2)} Wording variations among semantically equivalent answers can inflate pairwise semantic distances, leading methods such as SNNE~\cite{nguyen2025beyond} to overestimate uncertainty.
\textbf{(3)} Textual perturbations can dominate output variability, causing methods that rely on text paraphrasing or joint text-image perturbations (e.g., VL-Uncertainty~\cite{zhang2024vl} and C\&U~\cite{khan2024consistency}) to reflect prompt sensitivity rather than visual ambiguity.

First, Semantic Entropy~\cite{farquhar2024detecting} estimates uncertainty by sampling multiple outputs under stochastic decoding and computing the entropy of the answer distribution, where diverse answers correspond to high entropy and consistent answers to low entropy. While effective, this approach assumes that epistemic uncertainty manifests through decoding randomness. In VLMs, however, uncertainty can originate from the visual input which SE does not effectively capture (Fig.~\ref{fig:teaser}).
We show that visually biased images can induce visual embeddings that lead to highly confident predictions (Fig.~\ref{fig:h1}). Under such conditions, repeated stochastic decoding produces little output variation, rendering SE ineffective (Sec.~\ref{sec:confident_visual}).

Second, given the resulting answer variants, SE treats answers as categorical labels and estimates uncertainty by counting their frequencies, ignoring semantic distances between categories. Semantic Nearest Neighbor Entropy (SNNE)~\cite{nguyen2025beyond} addresses this by incorporating semantic distances through pairwise aggregation. However, SNNE operates directly on text outputs. Because \emph{language is discrete}, answers that share the same meaning but differ in wording may appear separated in embedding space. As a result, distance-based aggregation can  \emph{mistake wording variation for semantic disagreement} and inflate uncertainty (Sec.~\ref{sec:wording}).

\begin{figure}[t]
    \centering
    \includegraphics[width=\linewidth]{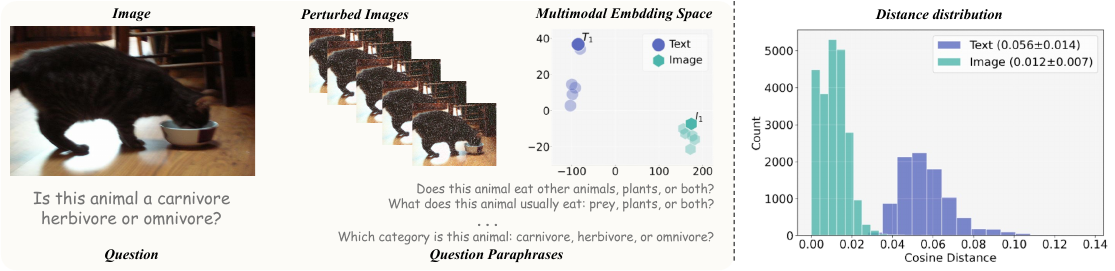}
    \caption{
\textbf{Text perturbations induce large semantic shifts.}
\textit{Left}: Given an input image and question, we generate perturbed images and textual paraphrases. 
In the multimodal embedding space, image perturbations form a tight local cluster around the original representation (gray dashed circle), while text paraphrases produce larger semantic shifts. 
\textit{Right}: Cosine distance distributions between perturbed samples and the original representation, measured on AOKVQA dataset~\cite{schwenk2022okvqa} using Qwen3-VL-Embedding-8B~\cite{li2026qwen3}. Text perturbations lead to substantially larger deviations than image perturbations (mean distance $0.056\pm0.014$ vs.\ $0.012\pm0.007$).  
}
    \label{fig:text_drive_large_deviation}
\end{figure}

Third, recent work~\cite{zhang2024vl,khan2024consistency} explores input-space perturbations through textual paraphrasing or joint text-image perturbations. 
As illustrated in Fig.~\ref{fig:text_drive_large_deviation}, due to \emph{prompt sensitivity}, text paraphrases can induce non-local, larger semantic shifts than visual perturbations, which typically produce only small, local changes. 
While such perturbations increase output diversity, the \emph{resulting variability is often dominated by textual changes} rather than visual evidence, with clustering is largely text-driven (Fig.~\ref{fig:h2_b}, Sec.~\ref{sec:text_dominate}).

To address these limitations, we propose \textbf{Visual Semantic Entropy (VSE)}, which perturbs the image to induce alternative answers and measures uncertainty through dispersion among semantic prototypes of the generated responses. \emph{First}, VSE only perturbs the input image, leaving the text query unchanged, to probe prediction instability that sampling alone fails to capture.
\emph{Second}, VSE clusters semantically similar answers into \emph{prototypes} so that wording variations do not inflate uncertainty. Uncertainty is measured as the frequency-weighted pairwise distance among prototypes.
\emph{Finally}, by perturbing only the image, VSE avoids modality confounding from textual perturbations and yields uncertainty that better reflects visual ambiguity.
Extensive experiments show that VSE is particularly effective on visually adversarial datasets, outperforming methods based on text perturbation and joint image-text perturbations.
In summary, our contributions are:
\begin{enumerate}
\item Our analysis reveals three failure modes of current VLMs uncertainty estimators: \textbf{(1)} suppressed output variation from overconfident visual embeddings; \textbf{(2)} inflated distances from wording variation; and \textbf{(3)} textual perturbations dominate output variability, causing uncertainty to reflect prompt sensitivity rather than visual ambiguity.
\item We introduce \textbf{Visual Semantic Entropy (VSE)}, which perturbs the image, clusters semantically similar answers, and quantifies uncertainty using weighted pairwise distances between prototype embeddings.
\item We present an extensive benchmark across 5 VLMs and 5 datasets, evaluating multiple uncertainty estimation methods and showing that VSE consistently yields more reliable uncertainty estimates.
\end{enumerate}
\section{Related Work}

\noindent\textbf{Uncertainty Estimation in VLMs} aims to quantify the reliability of model predictions, which is particularly important for visual question answering (VQA), where ambiguous visual evidence can lead to unreliable answers.

\textbf{Verbalized} methods prompt models to report confidence~\cite{groot2024overconfidence,pelucchi2023exploring,valdenegro2021find,xuan2025seeing}, while \textbf{logit-based} methods derive uncertainty from output probabilities~\cite{li2024reference,huangrevisiting}. Both do not explicitly account for visual ambiguity.

\textbf{Consistency-based} approaches, including SelfCheckGPT~\cite{manakul2023selfcheckgpt} and C\&U~\cite{khan2024consistency}, estimate uncertainty by measuring agreement between sampled responses and the original answer. C\&U additionally generates question paraphrases before probing agreement. However, when visual embeddings are confident, stochastic decoding can yield highly consistent outputs, causing these methods to underestimate uncertainty.

\textbf{Entropy-based} approaches estimate uncertainty from the semantic distribution of sampled answers. Semantic Entropy (SE)~\cite{farquhar2024detecting} computes entropy over outputs sampled under stochastic decoding. Later methods improve semantic representation, including SNNE~\cite{nguyen2025beyond} and KLE~\cite{nikitin2024kernel}, which aggregates pairwise distances across sampled answers. However, similar to consistency-based methods, these approaches rely on diversity in generated text under stochastic decoding and may fail to capture visual uncertainty when visual embeddings are confident, as shown in our analysis (Sec.~\ref{sec:confident_visual}). Moreover, pairwise distance aggregation can inflate uncertainty due to wording variations across answers(Sec.~\ref{sec:wording}).

\textbf{Input Perturbation.}
Recent work probes uncertainty through input perturbations. C\&U~\cite{khan2024consistency} perturbs the question, while VL-Uncertainty~\cite{zhang2024vl} perturbs both the image and the text. Our analysis shows that textual perturbations often dominate output variability due to prompt sensitivity, causing uncertainty estimates to reflect linguistic variation rather than visual ambiguity (Sec.~\ref{sec:text_dominate}).

Motivated by these limitations, we focus on uncertainty arising from visual ambiguity. Our design probes uncertainty through visual perturbations while avoiding text perturbations that introduce prompt sensitivity, and aggregates predictions at the semantic level to mitigate wording variations.
\section{Problem Analysis}

We analyze limitations of existing uncertainty estimators in VLMs and formalize two hypotheses and one proposition that guide our method.

\subsection{Problem Formulation}
A visual question instance consists of a question $q$ and an image $v$. 
Given $(q,v)$, a VLM produces an answer 
$a = \mathrm{VLM}(q,v)$.
Our goal is to estimate the reliability of this prediction.
Specifically, we seek an uncertainty estimator $U$ such that
$
\tilde{u} = U(q,v ; \mathrm{VLM})
$,
where larger uncertainty score $\tilde{u}$ indicates a higher likelihood that the generated answer $a$ is incorrect.
This is of high interests in scenarios where unreliable predictions must be filtered.

\subsection{Confident visual embeddings induce low semantic entropy}
\label{sec:confident_visual}

Semantic Entropy estimates uncertainty by sampling multiple outputs under stochastic decoding and computing the entropy of the resulting categorical answer distribution. This approach assumes that epistemic uncertainty manifests as variability induced by decoding randomness.
In VLMs, however, predictions are also conditioned on visual embeddings. When these embeddings are highly confident, stochastic decoding produces nearly identical outputs, leading to low SE even if the underlying image is ambiguous.

\paragraph{\textbf{Hypothesis 1}. Confident visual embeddings induce low semantic entropy.}
If the visual embedding yields a sharply peaked conditional answer distribution, decoding randomness alone cannot generate diverse outputs. As a result, SE may underestimate visual uncertainty.

\begin{figure}[!ht]
    \centering
    \includegraphics[width=0.5\linewidth]{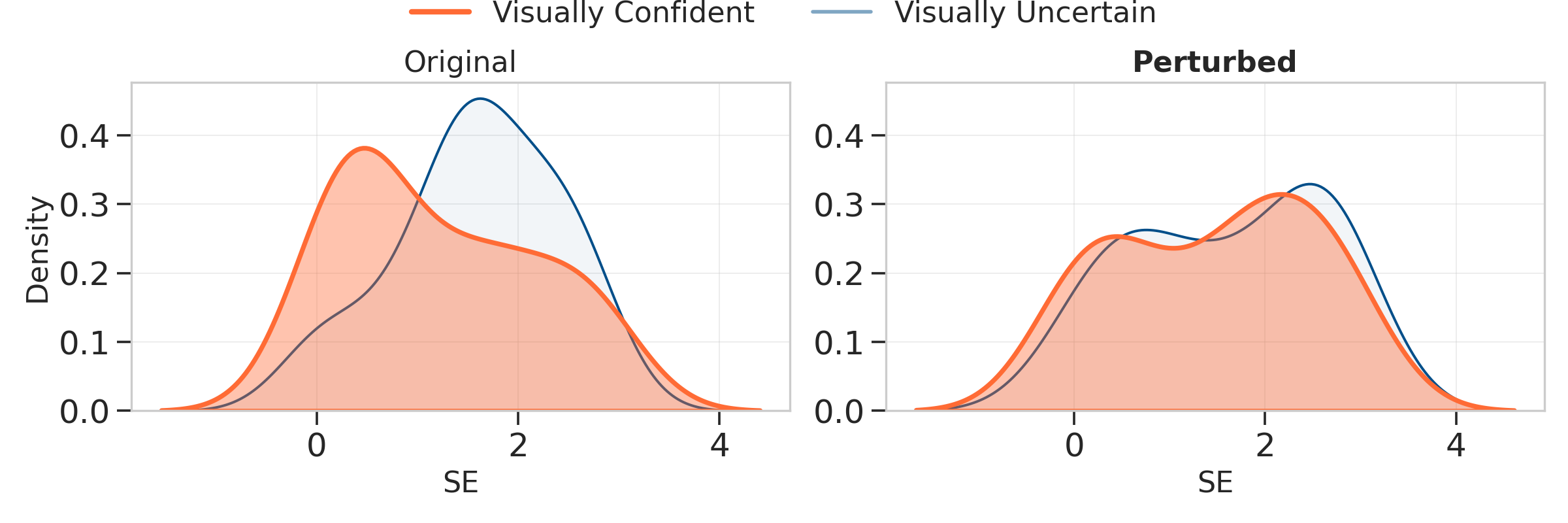}
    \caption{
    \textbf{Decoding-based Semantic Entropy underestimates visual ambiguity.}
    On the samples with incorrect answer, models are expected to express high uncertainty (SE). Using LogitLens~\cite{nostalgebraist2020interpreting} on the original image, we partition samples into \textcolor[RGB]{237, 80, 19}{Visually Confident} (low visual entropy) and \textcolor[RGB]{39,123,219}{Visually Uncertain} (high visual entropy).
    \textit{Left}: Under stochastic decoding, high SE appears only for \textcolor[RGB]{39,123,219}{Visually Uncertain} samples. \textcolor[RGB]{237, 80, 19}{Visually Confident} samples remain low despite the ambiguous input.
    \textit{Right}: After perturbing the image and aggregating predictions across perturbed views, we can induce high SE for \textcolor[RGB]{237, 80, 19}{Visually Confident} samples.
    Results are shown for Qwen2.5-VL on VILP dataset.
}
    \label{fig:h1}
\end{figure}

To test this hypothesis, we utilize the \textit{Adversarial} split of the VILP dataset~\cite{luo2024probing}, which consists of visually ambiguous images. We then filter samples that the VLM \emph{answers incorrectly} and quantify their visual confidence as follows: Let $\mathbf{v}_i\in\mathbb{R}^D$ denote the embedding of the $i$-th visual token at the final layer and $W\in\mathbb{R}^{\mathcal{V}\times D}$ the language modeling head with vocabulary size of $\mathcal{V}$ of the VLM.
Each visual token is projected into the vocabulary space using LogitLens ~\cite{nostalgebraist2020interpreting}:
\begin{equation}
\mathbf{z}_i = W \mathbf{v}_i,
\qquad
\Tilde{\mathbf{z}}_i = \mathrm{softmax}(\mathbf{z}_i).
\end{equation}
We compute the entropy of each visual token distribution and average across visual tokens to obtain the visual entropy $H_{\text{vis}}$:
\begin{equation}
H_{\text{vis}}(x)
=
\underset{i}{\mathrm{avg}}
\left(
-
\sum_{k=1}^{|\mathcal{V}|}
\Tilde{\mathbf{z}}_i(k)\log \Tilde{\mathbf{z}}_i(k)
\right).
\end{equation}
Lower visual entropy $H_{\text{vis}}$ indicates a more confident visual representation. Based on this metric, we partition samples into two groups: Visually confident (low visual entropy) and Visually uncertain (high visual entropy).
Importantly, this metric only reflects the model certainty \emph{in the visual input}, not the predicted answer.
We expect high SE (uncertainty) in both groups, since the model answers these samples incorrectly.
However, as shown in Fig.~\ref{fig:h1}-\textit{Left}, standard SE yields high uncertainty only for visually uncertain samples. Visually confident samples exhibit low SE despite the visual ambiguity, indicating that confident visual embeddings suppress output variability, yielding low semantic entropy.

We then perturb the input image to generate alternative local views while preserving semantic content.
As shown in Fig.~\ref{fig:h1}-\textit{Right}, perturbation induces higher semantic entropy for visually confident samples.
We note that the grouping is determined by calculating visual entropy $H_\text{vis}$ on the original image.
More details on experimental settings are included in the Supplementary.

\noindent\textbf{Discussion.}
These results demonstrate that decoding stochasticity alone fails under confident visual embeddings, while image perturbation induces high semantic entropy for visually ambiguous inputs. This motivates probing uncertainty in the visual input space.

\subsection{Text perturbation dominates output semantic shifts}
\label{sec:text_dominate}

Beyond perturbing visual inputs, recent work also perturbs the text query~\cite{zhang2024vl, khan2024consistency}. 
Unlike visual perturbations, which introduce small local changes around the same image representation, textual paraphrases can induce larger semantic shifts. 
Semantically equivalent sentences may occupy distant regions in embedding space due to the discrete nature of language. 
As a result, paraphrasing can substantially alter output semantics.
This motivates the following hypothesis.

\paragraph{\textbf{Hypothesis 2}.Text perturbations dominate output diversity under joint perturbation.}
We hypothesize that output diversity is driven primarily by textual variation rather than visual ambiguity. Then uncertainty measured under joint perturbation would mostly reflect \emph{prompt sensitivity}.

To test this, we analyze several VLMs on the \textit{Easy} and \textit{Adversarial} split of the VILP dataset.
For each image, we construct an $L \times M$ perturbation grid, where $L$ denotes the number of textual paraphrases and $M$ denotes the number of image perturbations. For each $(l, m)$ pair, we generate a model output and cluster the resulting $L \times M$ responses based on semantic similarity. This allows us to analyze how cluster membership depends on paraphrase identity or image perturbation.

\noindent\textbf{Purity analysis.}
To quantify the alignment between clusters and perturbation types, we compute cluster purity. 
For each cluster $c$, we define \emph{text purity} $P_T(c)$ and \emph{image purity} $P_I(c)$ as the fraction of samples in $c$ originating from the most frequent textual paraphrase and image perturbation. 
Let $n_c$ denote the size of cluster $c$, $n_c(l)$ the number of samples in $c$ generated from paraphrase $l$, and $n_c(m)$ the number generated from image perturbation $m$. We define
\begin{equation}
P_T(c) = \frac{\max_{l} n_c(l)}{n_c}, 
\quad
P_I(c) = \frac{\max_{m} n_c(m)}{n_c}.
\end{equation}

If $P_T(c) > P_I(c)$, most samples in cluster $c$ originate from a single paraphrase $l_{\max}$, indicating that cluster membership is primarily determined by the textual input rather than visual perturbations, and vice versa.
To obtain an overall text and image purity for each image across clusters, we compute the size-weighted purity: 
$P_T = \sum_{c} \frac{n_c}{L \times M} P_T(c)$ and 
$P_I = \sum_{c} \frac{n_c}{L \times M} P_I(c)$.

\begin{table}[t]
\caption{\textbf{Text ($P_T$) vs. Image ($P_I$) perturbations Purity} on VILP dataset for Qwen2.5-VL, Gemma3, Intern3.5-VL and LlaVA-NeXT.}
\centering
\small
\renewcommand{\arraystretch}{0.9}

\begin{tabular}{l
S S S
@{\hspace{18pt}}
S S S
@{\hspace{18pt}}
S S S
@{\hspace{18pt}}
S S S
}
\toprule
& \multicolumn{3}{l}{\textbf{Qwen2.5-VL}}
& \multicolumn{3}{l}{\textbf{Gemma3}}
& \multicolumn{3}{l}{\textbf{Intern3.5-VL}}
& \multicolumn{3}{l}{\textbf{LlaVA-NeXT}}\\

& \emph{Easy} & \emph{Adv.} & \emph{All}
& \emph{Easy} & \emph{Adv.} & \emph{All}
& \emph{Easy} & \emph{Adv.} & \emph{All}
& \emph{Easy} & \emph{Adv.} & \emph{All} \\
\midrule
$P_T$(\%)
& 51.9 & 55.6 & 54.3
& 35.6 & 38.3 & 37.4
& 46.7 & 48.4 & 47.8
& 54.5 & 53.9 & 54.1\\

$P_I$(\%)
& 36.4 & 40.1 & 38.8
& 30.4 & 36.8 & 34.6
& 37.8 & 41.4 & 40.2
& 41.0 & 44.8 & 43.4\\
\bottomrule
\end{tabular}

\label{tab:purity}
\end{table}

\begin{figure}[!]
    \vspace{-1em}
    \centering
    \includegraphics[width=0.9\linewidth]{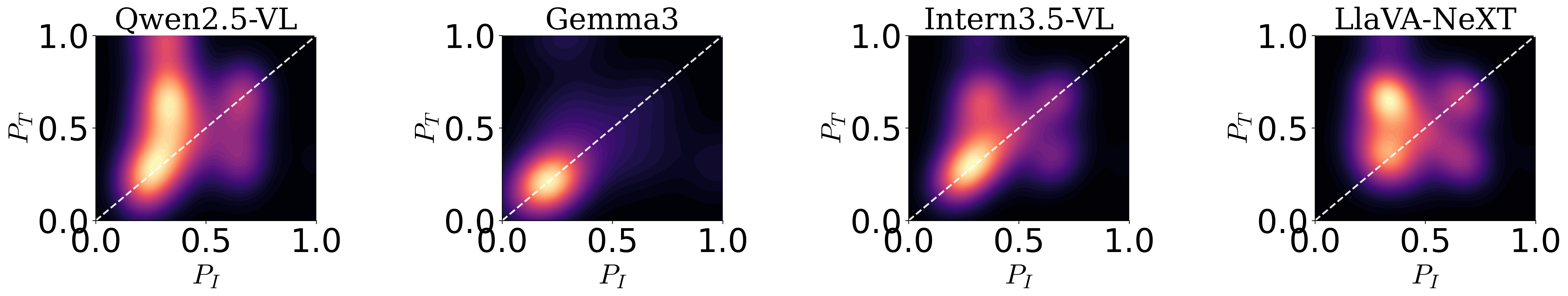}
    \caption{\textbf{Kernel density of text $P_T$ and image purity $P_I$ across models.}. The diagonal $P_T = P_I$ separates text- from image-dominant regions. Density mass lying predominantly above the diagonal indicates that clustering is primarily text-driven.}
    \vspace{-1em}
    \label{fig:h2_c}
    \vspace{-1em}
\end{figure}

Overall, text purity consistently exceeds image purity (Tab.~\ref{tab:purity}), and the gap widens on the \emph{Easy} split. Furthermore, we estimate the joint density of $(P_T, P_I)$ across images using kernel smoothing and visualize it as a filled contour plot  (Fig.~\ref{fig:h2_c}). The diagonal $P_T = P_I$ separates text-dominant (above) from image-dominant regions (below). We observe that the density mass concentrates predominantly above this line, meaning that for most images, cluster assignments align more strongly with paraphrase identity than with visual perturbations.
This pattern shows that variability under joint perturbation is largely text-driven rather than visual ambiguity. Additional visualizations and details are in Supp.

\begin{figure}[H]
    \centering
    
    \begin{subfigure}{0.48\textwidth}
        \centering
        \includegraphics[width=\linewidth]{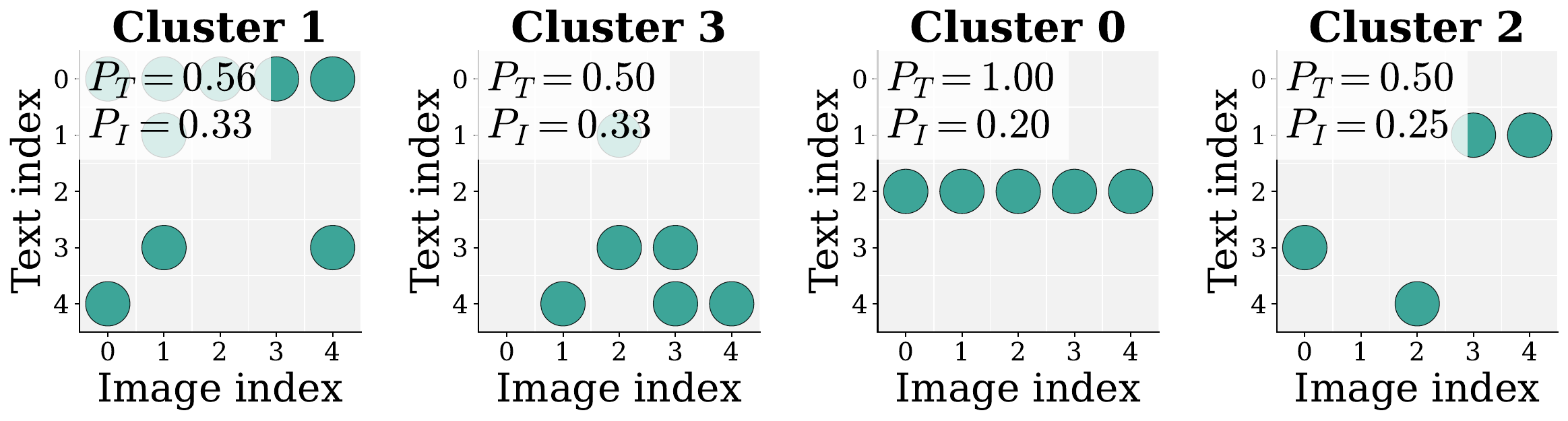}
        \label{fig:sub1}
    \end{subfigure}
    \hfill
    \begin{subfigure}{0.48\textwidth}
        \centering
        \includegraphics[width=\linewidth]{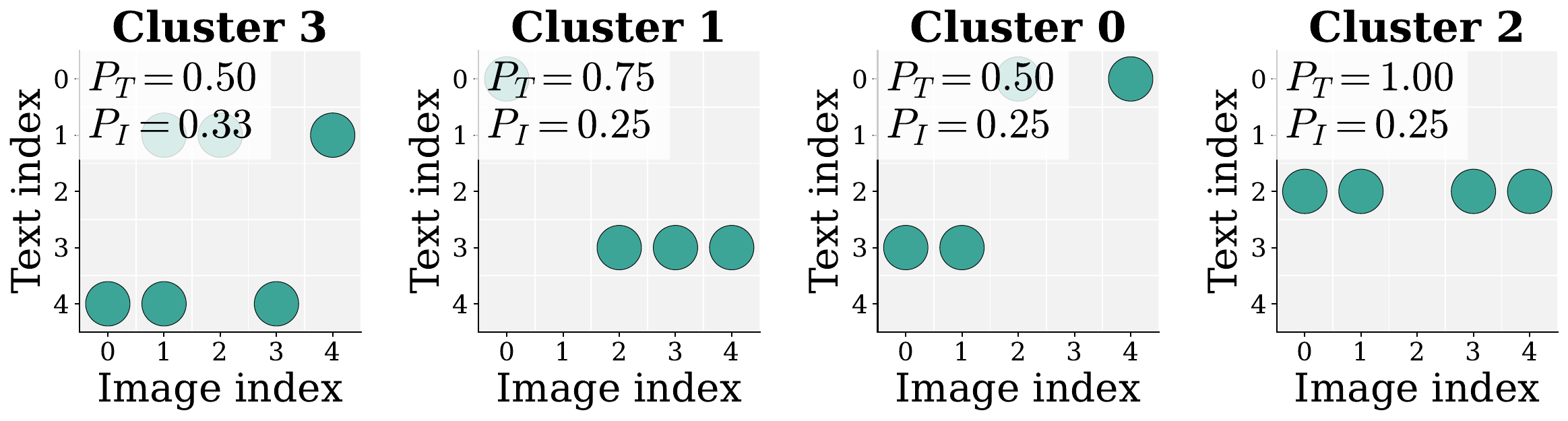}
        \label{fig:sub2}
    \end{subfigure}
    
    \vspace{-0.1cm}
    
    \begin{subfigure}{0.48\textwidth}
        \centering
        \includegraphics[width=\linewidth]{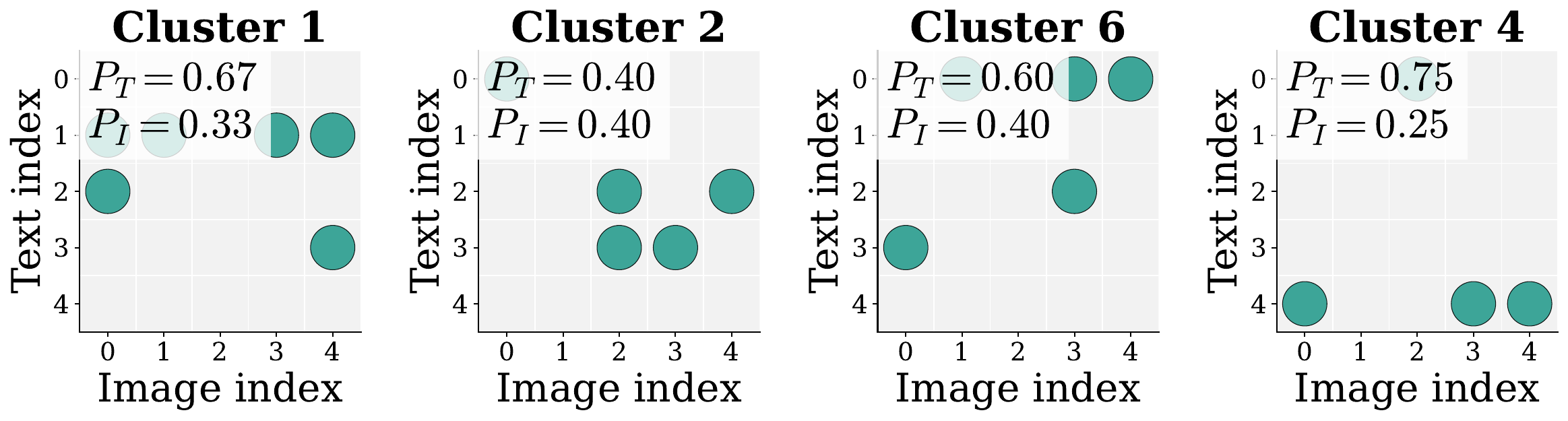}
        \label{fig:sub3}
    \end{subfigure}
    \hfill
    \begin{subfigure}{0.48\textwidth}
        \centering
        \includegraphics[width=\linewidth]{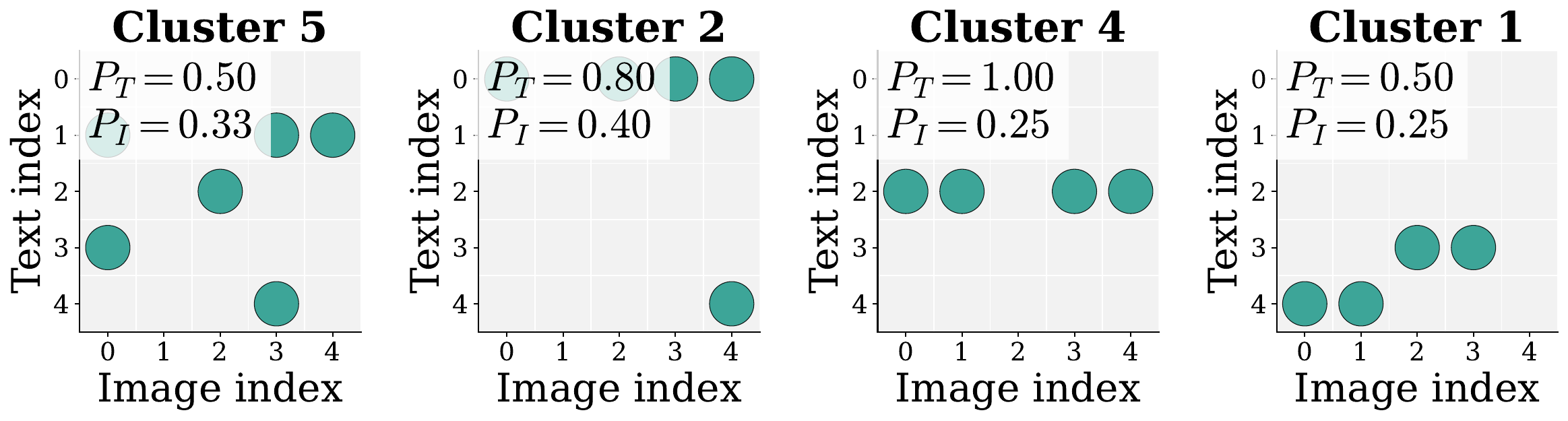}
        \label{fig:sub4}
    \end{subfigure}
    \vspace{-2em}
    \caption{\textbf{Cluster occupancy map.}
Each panel shows cluster assignments on the $L \times M$ perturbation grid  of a random sample (rows: textual paraphrases; columns: image perturbations). Horizontal stripes indicate invariance across image perturbations, showing that clustering is dominant by textual paraphrase. We provide more results in Supp.}

    \label{fig:h2_b}
\end{figure}

To qualitatively examine this effect, we visualize cluster occupancy on the $L \times M$ perturbation grid. Rows correspond to textual paraphrases and columns to image perturbations. Under joint perturbation, we observe \emph{horizontal stripe patterns} in the occupancy maps (Fig.~\ref{fig:h2_b}), indicating that cluster membership is invariant across image perturbations but consistent within a given paraphrase. This pattern confirms that output are primarily conditioned on the textual input.

\noindent\textbf{Discussion.}
The quantitative purity analysis and qualitative examples support Hypothesis~2: under joint image-text perturbation, output diversity is largely driven by textual variation rather than visual ambiguity. Our finding aligns with Image-DPO~\cite{luo2024probing}, where the authors perturb only the image while keeping the question fixed to dismiss model over-reliance on language. In uncertainty estimation, prompt sensitivity leads paraphrasing to induce semantic shifts that inflate uncertainty without reflecting ambiguity in visual information. 
Hence, restricting perturbations to the visual domain avoids artificial prompt sensitivity and yields more faithful visual uncertainty.

\subsection{Linguistic variations inflates distance-based uncertainty}
\label{sec:wording}

Other approaches estimate uncertainty by measuring semantic distances among sampled outputs~\cite{nguyen2025beyond, nikitin2024kernel}. 
However, because answers are discrete text, semantically equivalent responses with different wording can be mapped to distant points in embedding space. 
As a result, distance-based aggregation may yield inflated uncertainty even when there is no true semantic disagreement.

Formally, given $n$ sampling iterations, the model generates answer variants $\{a_i\}^n$. 
A general distance-based estimator computes uncertainty as the average pairwise distance:
\begin{equation}
U(q)=\frac{1}{n(n-1)}\sum \mathrm{d}(a,a'),
\label{eq:pairwise}
\end{equation}
where $\mathrm{d}(\cdot,\cdot)$ is a semantic distance function.

\paragraph{\textbf{Proposition 1}. Distance inflation under linguistic variations.}
Suppose all sampled answers express the same semantic meaning. 
If wording differences induce non-zero embedding distances, 
$\mathrm{d}(a,a') > 0$ for some $a \neq a'$, 
then $U(q) > 0$ despite the absence of semantic disagreement.

\paragraph{Intra- vs. Inter-cluster effect.}
To separate meaning-level disagreement from wording variation, 
consider partitioning the answers into semantic clusters, where each 
cluster groups answers that convey the same meaning. Let $c$ denote 
the cluster assignment of $a$.
For a distance metric $\mathrm{d}(\cdot,\cdot)$, pairwise aggregation can be 
decomposed as:

\begin{align}
U(q)
=
\sum_{k=1}^K
\,\mathbb{E}\!\left[\mathrm{d}(a,a') \mid c=c'=k\right]
 +
\sum_{k\neq k'} 
\,\mathbb{E}\!\left[\mathrm{d}(a,a') \mid c=k, c'=k'\right].
\label{eq:decomposition}
\end{align}
The first term captures variation within clusters, which mainly reflects 
wording differences.
The second term captures 
variation across clusters, corresponding to genuine semantic disagreement. 
Because distance-based aggregation sums both terms, uncertainty is 
influenced not only by disagreement between meanings but also by wording variation within each meaning. 
Therefore, uncertainty estimation should first group semantically equivalent answers and then measure distances across groups rather than 
between individual text outputs.

\section{Method}

The analysis above shows three limitations of current uncertainty estimators: (1) decoding stochasticity underestimates uncertainty when visual embeddings are overly confident, (2) text perturbations can dominate output variability under joint image-text perturbation, and (3) linguistic variations inflate distance-based estimators. 
These observations suggest that VQA uncertainty estimation should probe ambiguity in the visual input space,  avoid textual perturbations that introduce large semantic shifts, and  aggregate outputs at the semantic level.

\begin{figure}
    \centering
    \includegraphics[width=\linewidth]{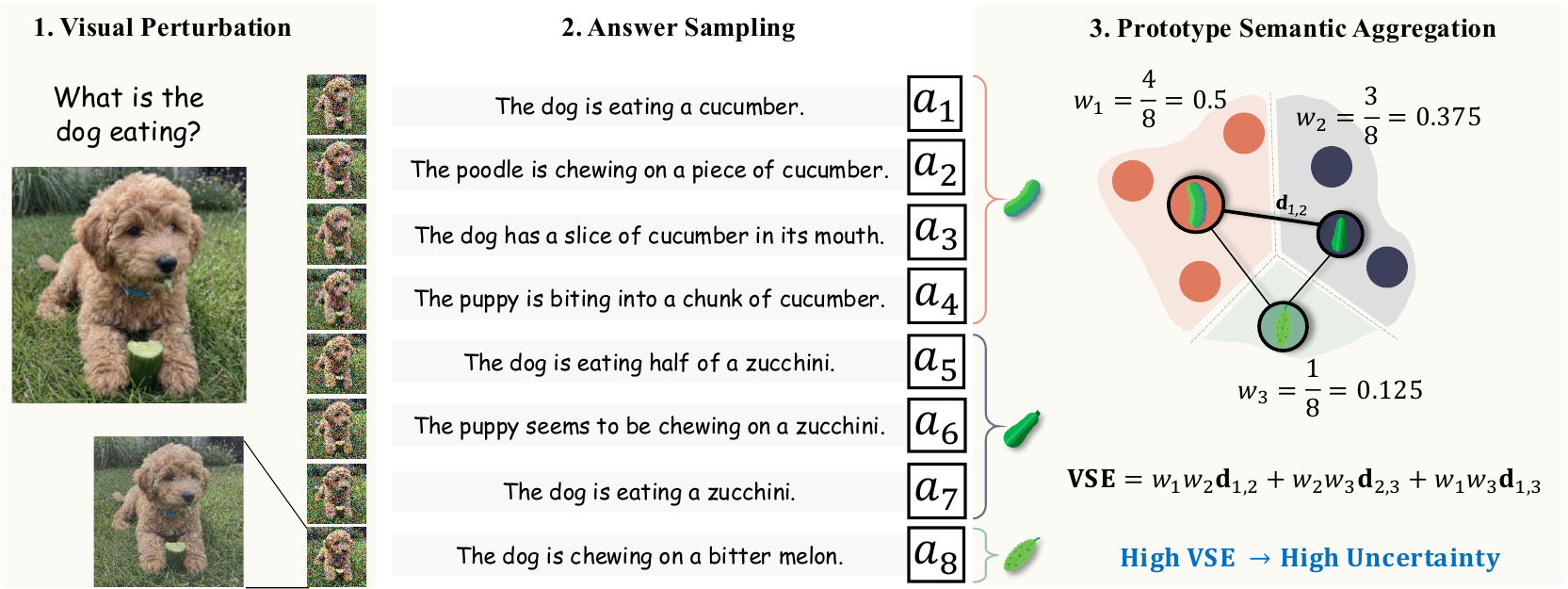}
    \caption{
    \textbf{Visual Semantic Entropy for VQA Uncertainty Estimation.}
    (1) We perturb the input image to generate local visual variants.
    (2) The VLM produces multiple answer samples across perturbed views. (3) We cluster semantically equivalent answers, select a representative prototype for each cluster, and compute uncertainty as the mass-weighted dispersion among prototype embeddings.}
    \label{fig:method}
\end{figure}

Guided by these principles, we propose \textbf{Visual Semantic Entropy (VSE)}, a method that perturbs the image to elicit alternative visual interpretations (Sec.~\ref{sec:perturb}), clusters semantically equivalent answers into prototypes, and quantifies uncertainty as the weighted dispersion among prototype embeddings (Sec.~\ref{sec:protosem}). 
An overview of VSE is shown in Fig.~\ref{fig:method}.

\subsection{Visual Perturbation}
\label{sec:perturb}

To probe visual epistemic uncertainty, we perturb the input image while keeping the question fixed. 
Unlike textual paraphrasing, which may induce non-local semantic shifts, visual perturbations introduce controlled variations around the original image, enabling us to explore local visual variants.

Let $\mathcal{T}$ denote a visual perturbation operator. 
Given an input image $v$, we generate $M$ perturbed views:
\begin{align}
   v_m = \mathcal{T}(v; \xi_m), 
    \quad m = 1,\dots,M, 
\end{align}
where $\xi_m$ parameterizes the perturbation. 
The operator $\mathcal{T}$ is designed to preserve high-level semantic content while inducing local variations in pixel space.
Each perturbed image $v_m$ is paired with the same question $q$ and fed into the VLM to obtain answer samples. 
Variability across these perturbed views $v_m$ serves as a probe of visual uncertainty.

\noindent\textbf{Discussion.} Prior work~\cite{zhang2024vl} progressively increases perturbation magnitude to generate samples at multiple noise levels. 
In contrast, we fix a small perturbation scale for two reasons.
First, uncertainty should quantify ambiguity conditioned on the original input $(q,v)$. 
Progressively increasing perturbations magnitude may move samples outside the local neighborhood of $v$, so answer variation reflects changes to the input rather than visual ambiguity.
Second, sampling-based estimators aggregate samples uniformly. 
Mixing perturbation magnitudes inappropriately assigns equal weight to local and non-local, large deviations.
By using a single fixed scale, the measured dispersion consistently reflects variability within a local neighborhood.

\subsection{Prototype Semantic Aggregation}
\label{sec:protosem}

Given $M$ perturbed image variants $\{v_m\}^M$ and the question $q$, the VLM produces a corresponding set of answer samples 
$\{a_m\}^M$.
These samples may differ in wording or semantic meaning.
We then perform Prototype Semantic Aggregation (\textbf{ProtoSem}) as follows:

\textit{Clustering.}
We cluster the answer variants $\{a_m\}$  into $K$ groups 
$\{c_k\}_{k=1}^K$
based on semantic distance base on a function $\mathrm{d}(\cdot,\cdot)$.

\textit{Prototype selection.}
For each cluster $c_k$, we select a prototype answer
$p_k$
that minimizes the average distance to other members in the same cluster:
\begin{equation}
p_k
=
\arg\min_{a \in c_k}
\sum_{a' \in c_k}
\mathrm{d}(a, a').
\end{equation}
This prototype serves as the representative semantic meaning of cluster $k$.

\textit{Prototype dispersion.}
Let $w_k=\frac{|c_k|}{M}$ denote the empirical probability of cluster $k$.
We define uncertainty as the expected semantic distance between two independently drawn cluster prototypes:
\begin{equation}
\Tilde{u}
=
U(q,v; \text{VLM})
=
\mathbb{E}_{k \sim w,\; k' \sim w}\!\left[\mathrm{d}(p_k,p_{k'})\right]
=
\sum_{k \neq k'} w_k w_{k'} \, \mathrm{d}(p_k,p_{k'}).
\label{eq:vpd}
\end{equation}
This measures disagreement among supported semantic meanings: 
semantic variations are consolidated within each cluster and represented by a single prototype $p_k$, and each prototype is weighted by its mass $w_k$. 
Uncertainty increases only when \emph{multiple high-mass prototypes are separated}.

\noindent\textbf{Discussion.}
SE~\cite{farquhar2024detecting} treats answers as discrete categories and ignores semantic proximity, whereas SNNE~\cite{nguyen2025beyond} measures pairwise distances and is sensitive to linguistic variation. 
VSE instead groups semantically equivalent outputs and compute weighted dispersion over cluster prototypes, eliminating intra-cluster wording inflation while preserving semantic disagreement across clusters.
\section{Experiments}
\subsection{Setup}

\noindent\textbf{Metrics.}
Following prior work~\cite{farquhar2024detecting,nguyen2025beyond}, we evaluate uncertainty estimation using the Area Under the ROC Curve (AUC). Each prediction is labeled as correct or incorrect, and the predicted uncertainty score is used to distinguish between them. AUC measures how well uncertainty scores rank incorrect predictions above correct ones across different thresholds. A higher AUC indicates better alignment between predicted uncertainty and model reliability.

\noindent\textbf{Datasets.}
We evaluate our method on five benchmarks covering knowledge-based VQA, multimodal reasoning: OKVQA~\cite{marino2019ok}, AOKVQA~\cite{schwenk2022okvqa},MMVet~\cite{yu2024mm}; and adversarial datasets to expose visual bias: VILP~\cite{luo2024probing}, VLM-are-biased~\cite{vo2025vision}.
We detail on the dataset and their splits in the Supplementary material.

\noindent\textbf{Other Baselines.}
We conduct an extensive benchmark to compare \textbf{VSE} against these uncertainty estimation approaches:
\emph{Verbalized}: Verbalized Uncertainty~\cite{groot2024overconfidence}
.
\emph{Logit-based}: Confidence measures derived from output probabilities, including AvgEnt and MaxEnt (token entropy), and AvgProb and MaxProb (token probability)~\cite{li2024reference}.
\emph{Consistency-based}: 
SelfCheckGPT~\cite{manakul2023selfcheckgpt} and C\&U~\cite{khan2024consistency}.
\emph{Entropy-based}: 
Semantic Entropy (SE)~\cite{farquhar2024detecting}, Semantic Nearest Neighbor Entropy (SNNE)~\cite{nguyen2025beyond}, Kernel Language Entropy (KLE)~\cite{nikitin2024kernel}, and VL-Uncertainty~\cite{zhang2024vl}. Our method, \textbf{VSE}, also belongs to this category.

\noindent\textbf{Models.}
We evaluate all approaches on five vision-language models that cover a range of architecture: Qwen2.5-VL-7B~\cite{bai2025qwen25vl}, Gemma3-4B~\cite{gemma2025gemma3}, Intern3.5-VL-8B~\cite{wang2025internvl3}, LLaVA-NeXt-8B~\cite{liu2024llavanext}, and Qwen3-VL-8B~\cite{bai2025qwen3}. 

\noindent\textbf{Implementation Details.}
To perturb the image, we add Gaussian noise with standard deviation $\sigma=20$, introducing small visual variations while preserving semantic content. 
The initial answer $a$ is generated using greedy decoding ($T=0.0$). 
After obtaining this initial answer, we sample $M=10$ additional responses with decoding temperature $T=1.0$, following the setup of SE~\cite{farquhar2024detecting}. 
For answer clustering, we use the DeBERTa-v2-xlarge-mnli~\cite{he2020deberta} as the semantic distance function $\mathrm{d}(\cdot,\cdot)$ and apply hierarchical clustering~\cite{nielsen2016hierarchical}. 
Additional hyperparameter analysis can be found in supplementary material.

\subsection{Results}

\noindent\textbf{Qualitative Results.}
We report results on AOKVQA in Tab.~\ref{tab:aokvqa_results}. Entropy-based methods perform best overall, while logit-based approaches perform poorly across all models. Verbalized uncertainty performs competitively only on Qwen3-VL. Consistency-based methods also show strong performance, particularly C\&U with text perturbations. Among entropy-based methods, \textbf{VSE} consistently achieves the best results across all models, reaching AUC scores of 0.783, 0.778, 0.724, 0.798, and 0.792 on Qwen2.5-VL, Gemma3, LLaVA-NeXT, Qwen3-VL, and Intern3.5-VL, and outperforming VL-Uncertainty by up to 9.2\% AUC.
The same trend holds on MMVet and OKVQA (Tabs.~\ref{tab:mmvet_results} and~\ref{tab:okvqa_results}). VSE achieves the best performance across all models, improving over the strongest baselines such as SE (e.g., 0.767 vs.\ 0.758 on Qwen2.5-VL in OKVQA) and reaching AUC scores of 0.781 and 0.778 on MMVet for Qwen2.5-VL and Gemma3, respectively.
We observe that SE and SNNE perform well on MMVet, suggesting that repeated sampling better captures uncertainty in this benchmark, which primarily evaluates reasoning ability rather than visual ambiguity.

\noindent\textbf{Qualitative Results.}
We provide qualitative result in Fig.~\ref{fig:qualitative_main}. More results are provided in Supplementary.

\begin{figure}[!]
    \centering
    \includegraphics[width=\linewidth]{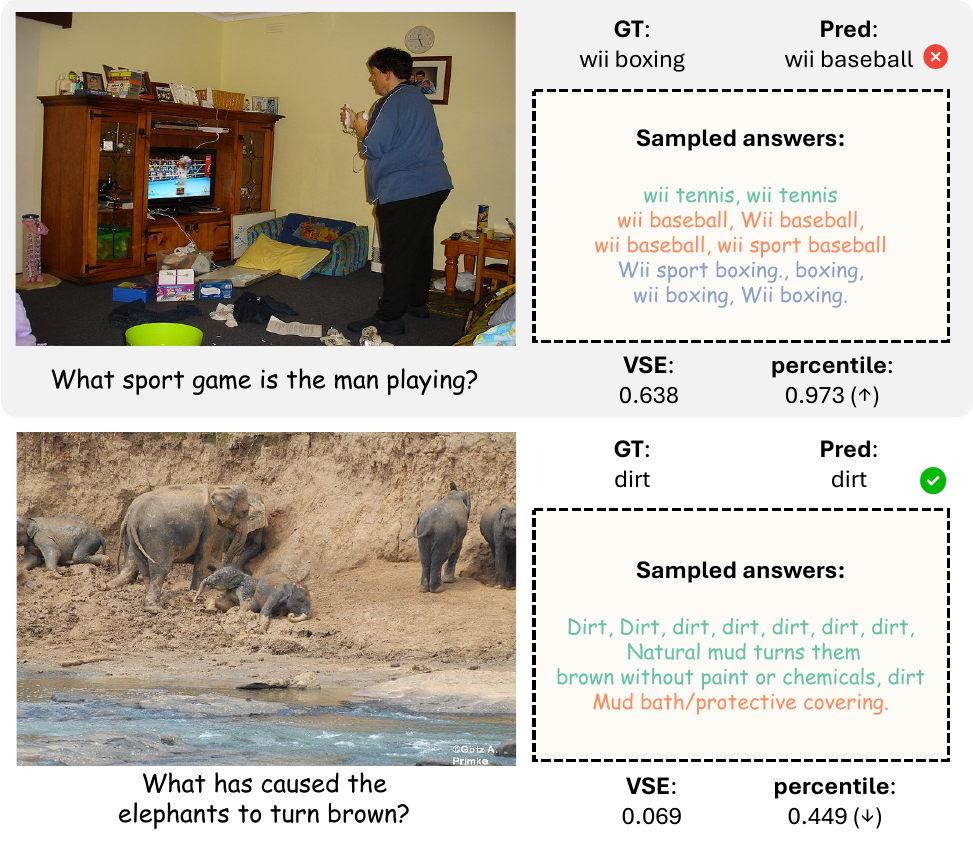}
    \caption{
\textbf{Qualitative Result of Visual Semantic Entropy.}
\textit{Top}: VSE is high, showing high uncertainty for wrong prediction.
\textit{Bottom}: VSE is low, showing more certainty for correct prediction.
Results are shown for Qwen2.5-VL on AOKVQA.
}
    \label{fig:qualitative_main}
\end{figure}

\begin{table}[!]
\caption{\textbf{AOKVQA results.} AUC scores of uncertainty estimation methods across Qwen2.5-VL, Gemma3, LLaVA-NeXT, Qwen3-VL, and Intern3.5-VL. Higher is better.}
\label{tab:aokvqa_results}
\centering
\footnotesize
\renewcommand{\arraystretch}{1.05}
\resizebox{0.9\linewidth}{!}{
\begin{tabular}{l|l|c|ccccc}
\toprule
\textbf{Category} & \textbf{Method} & \textbf{Venue} & \textbf{Qwen2.5} & \textbf{Gemma3} & \textbf{LLaVA} & \textbf{Qwen3} & \textbf{Intern3.5} \\
\midrule

\multirow{1}{*}{\textit{Verbalized}}
& Verb-U~\cite{groot2024overconfidence} & NAACL'24   & 0.635 & 0.654 & 0.580 & 0.700 & 0.630 \\

\midrule
\multirow{4}{*}{\textit{Logit}}
& AvgEnt~\cite{li2024reference} & EMNLP'24  & 0.625 & 0.530 & 0.635 & 0.581 & 0.613 \\
& MaxEnt~\cite{li2024reference} & EMNLP'24  & 0.596 & 0.529 & 0.638 & 0.578 & 0.632 \\
& AvgProb~\cite{li2024reference} & EMNLP'24  & 0.612 & 0.508 & 0.634 & 0.556 & 0.593 \\
& MaxProb~\cite{li2024reference} & EMNLP'24  & 0.600 & 0.561 & 0.522 & 0.541 & 0.512 \\

\midrule
\multirow{3}{*}{\textit{Consistency}}
& SCG-NLI~\cite{manakul2023selfcheckgpt} & EMNLP'23  & 0.575 & 0.576 & 0.579 & 0.591 & 0.618 \\
& SCG-Pr~\cite{manakul2023selfcheckgpt} & EMNLP'23  & 0.734 & 0.701 & 0.608 & 0.708 & 0.730 \\
& C\&U~\cite{khan2024consistency} & CVPR'24  & 0.731 & 0.712 & 0.602 & 0.720 & 0.734 \\

\midrule
\multirow{5}{*}{\textit{Entropy}}
& SE~\cite{farquhar2024detecting} & ICLR'23   & 0.702 & \underline{0.732} & 0.622 & 0.746 & 0.775 \\
& SNNE~\cite{nguyen2025beyond} & ACL'25  & 0.744 & 0.700 & \underline{0.651} & 0.747 & 0.745 \\
& KLE~\cite{nikitin2024kernel} & NIPS'25  & \underline{0.774} & 0.721 & 0.541 & \underline{0.772} & 0.764 \\
& VL-U~\cite{zhang2024vl} & arxiv  & 0.756 & 0.715 & 0.616 & 0.706 & \underline{0.781} \\
\rowcolor{green!7}
& \textbf{VSE} & \textbf{Ours}  & \textbf{0.783} & \textbf{0.778} & \textbf{0.724} & \textbf{0.798} & \textbf{0.792} \\

\bottomrule
\end{tabular}
}

\end{table}

\begin{table}[!]
\caption{\textbf{MMVet results.} AUC scores of uncertainty estimation methods across Qwen2.5-VL, Gemma3. Higher is better.}
\label{tab:mmvet_results}
\centering
\footnotesize
\resizebox{0.95\linewidth}{!}{
\begin{tabular}{lccccccccc}
\toprule
Model 
& Verb-U~\cite{groot2024overconfidence}
& A-E~\cite{li2024reference}
& A-P~\cite{li2024reference}
& SCG~\cite{manakul2023selfcheckgpt}
& C\&U~\cite{khan2024consistency}
& SE~\cite{farquhar2024detecting}
& SNNE~\cite{nguyen2025beyond}
& VL-U~\cite{zhang2024vl}
& \textbf{VSE} \\
\midrule
\textbf{Qwen2.5-VL} & 0.595  & 0.687 & 0.678 & 0.718 & 0.587 & 0.716 & \underline{0.758} & 0.659 & \textbf{0.781} \\
\textbf{Gemma3}  & 0.598  & 0.607 & 0.601 & 0.625 & 0.646 & 0.722 & \underline{0.740} & 0.693 & \textbf{0.778} \\
\bottomrule
\end{tabular}
}
\end{table}

\begin{table}[t]
\caption{\textbf{OKVQA results.} AUC scores of uncertainty estimation methods across Qwen2.5-VL, Gemma3, LLaVA-NeXT, Qwen3-VL. Higher is better.}
\label{tab:okvqa_results}
\centering
\footnotesize
\renewcommand{\arraystretch}{1.05}
\resizebox{0.8\linewidth}{!}{
\begin{tabular}{l|l|c|cccc}
\toprule
\textbf{Category} & \textbf{Method} & \textbf{Venue} & \textbf{Qwen2.5} & \textbf{Gemma3} & \textbf{LLaVA} & \textbf{Qwen3} \\
\midrule

\multirow{1}{*}{\textit{Verbalized}}
& Verb-U~\cite{groot2024overconfidence} & NAACL'24   & 0.663 & 0.647 & 0.532 & 0.593  \\

\midrule
\multirow{4}{*}{\textit{Logit}}
& AvgEnt~\cite{li2024reference} & EMNLP'24  & 0.690 & 0.577 & \underline{0.729} & 0.703 \\
& MaxEnt~\cite{li2024reference} & EMNLP'24  & 0.714 & 0.567 & 0.724 & 0.715 \\
& AvgProb~\cite{li2024reference} & EMNLP'24  & 0.653 & 0.570 & 0.705 & 0.679  \\
& MaxProb~\cite{li2024reference} & EMNLP'24  & 0.519 & 0.555 & 0.542 & 0.565 \\

\midrule
\multirow{3}{*}{\textit{Consistency}}
& SCG-NLI~\cite{manakul2023selfcheckgpt} & EMNLP'23  & 0.714 & 0.584 & 0.658 & 0.597  \\
& SCG-Pr~\cite{manakul2023selfcheckgpt} & EMNLP'23  & \textbf{0.767} & 0.657 & 0.675 & 0.657 \\
& C\&U~\cite{khan2024consistency} & CVPR'24  & {0.742} & {0.678} & {0.718} & {0.693}  \\

\midrule
\multirow{5}{*}{\textit{Entropy}}
& SE~\cite{farquhar2024detecting} & ICLR'23  &\underline{ 0.758} & \underline{0.703} & 0.685 & \underline{0.790}  \\
& SNNE~\cite{nguyen2025beyond} & ACL'25  & 0.751 & 0.667 & 0.698 & 0.711  \\
& KLE~\cite{nikitin2024kernel} & NIPS'25  & 0.744 & 0.673 & 0.672 & 0.724 \\
& VL-U~\cite{zhang2024vl} & arxiv  & {0.731} & 0.702 & {0.695} & {0.731} \\
\rowcolor{green!7}
& \textbf{VSE} & \textbf{Ours}  & \textbf{0.767} & \textbf{0.715} & \textbf{0.749} & \textbf{0.799} \\

\bottomrule
\end{tabular}
}
\end{table}

\subsection{Ablation Study}

\noindent\textbf{VSE reflects visual ambiguity more faithfully.}
We evaluate uncertainty estimation on the visually adversarial datasets VILP and VLM-are-biased (Tab.~\ref{tab:vilp_vlmbiased}), where images are intentionally designed to be highly ambiguous to probe model biases and test reliance on visual evidence. 
VSE consistently achieves the best performance across all settings, improving over the strongest baseline by large margins (e.g., $0.650$ vs.\ $0.535$ on VILP with Qwen2.5-VL and $0.826$ vs.\ $0.783$ on VLM-are-biased). In contrast, several existing methods degrade substantially on these datasets, suggesting that they are sensitive to language biases or decoding artifacts. These results highlight that explicitly probing visual variations help to capture uncertainty arising from ambiguous visual inputs.

\begin{table}
\caption{AUC scores of uncertainty estimation methods on the VILP and VLM-are-biased benchmarks.}
\label{tab:vilp_vlmbiased}
\centering
\footnotesize
\renewcommand{\arraystretch}{1.15}
\begin{tabular}{llcccccc}
\toprule
\textbf{Dataset} & \textbf{Model} &Verb-U~\cite{groot2024overconfidence} & A-E~\cite{li2024reference} & SE~\cite{farquhar2024detecting} & C\&U~\cite{khan2024consistency} & VL-U~\cite{zhang2024vl} & \textbf{VSE} \\
\midrule
\multirow{2}{*}{VILP}
 & \textbf{Qwen2.5-VL} & 0.521 & 0.507 & \underline{0.535} & 0.515 & 0.489 & \textbf{0.650} \\
 & \textbf{Gemma3}     & 0.503 & 0.528 & \underline{0.660} & 0.540 & 0.501 & \textbf{0.665} \\
\midrule
\multirow{2}{*}{VLM-are-biased}
 & \textbf{Qwen2.5-VL} & 0.722 & 0.537 & 0.728 & 0.697 & \underline{0.783} & \textbf{0.826} \\
 & \textbf{Gemma3}     & 0.600 & 0.694 & 0.700 & 0.737 & \underline{0.758} & \textbf{0.776} \\
\bottomrule
\end{tabular}

\end{table}

\noindent\textbf{Effect of Visual Perturbation and Prototype Aggregation.}
Tab.~\ref{tab:perturbation_entropy} evaluates two components of our approach. For SE and SNNE, we report the original estimator (\emph{Base}) and a variant where we add visual perturbation ($+\mathcal{T}$). We also evaluate \textbf{ProtoSem}, our prototype semantic aggregation strategy, both with and without visual perturbation.

(1) Comparing \emph{Base} and $+\mathcal{T}$ shows consistent improvements for SE and SNNE across all models, with gains up to $+9.5\%$ AUC, demonstrating that visual perturbations are beneficial for uncertainty estimation.

(2) Comparing methods across $+\mathcal{T}$ columns (with visual perturbation), \textbf{ProtoSem} consistently achieves the best performance, demonstrating that prototype-based semantic aggregation better captures uncertainty than entropy computed over sampled answers, as used in SE and SNNE.

These results justify the core design of VSE, combining visual perturbation with prototype-based semantic aggregation to improve uncertainty estimation.

\begin{table}[!]
\caption{\textbf{Effect of visual perturbation ($+\mathcal{T}$) and Prototype Semantic Aggregation (ProtoSem)} on entropy-based uncertainty estimation methods on the AOKVQA dataset. The $\text{ProtoSem}+\mathcal{T}$ combination is our proposed \textbf{VSE}.}
\label{tab:perturbation_entropy}
\centering
\footnotesize
\renewcommand{\arraystretch}{1.15}
\begin{tabular}{l@{\hspace{10pt}}ccc@{\hspace{10pt}}ccc@{\hspace{10pt}}ccc}
\toprule
 & \multicolumn{3}{c}{SE~\cite{farquhar2024detecting}} 
 & \multicolumn{3}{c}{SNNE~\cite{nguyen2025beyond}} 
 & \multicolumn{3}{c}{\textbf{ProtoSem}} \\
\cmidrule(r{10pt}){2-4}
\cmidrule(r{10pt}){5-7}
\cmidrule{8-10}
Model & Base & $+\mathcal{T}$ & $\Delta$ 
      & Base & $+\mathcal{T}$ & $\Delta$ 
      & Base & $+\mathcal{T}_{(\textbf{VSE})}$ & $\Delta$ \\
\midrule

\textbf{Qwen2.5-VL} & 0.702 & 0.761 & +0.059 & 0.700 & 0.720 & +0.020 & 0.711 & 0.783 & +0.072  \\
\textbf{Gemma3}     & 0.732 & 0.775 & +0.043 & 0.651 & 0.746 & +0.095 &  0.745 & 0.778 & +0.033  \\
\textbf{Qwen3-VL}   & 0.747 & 0.788 & +0.041 & 0.747 & 0.757 & +0.010 & 0.762 & 0.798  & +0.036  \\
\bottomrule
\end{tabular}
\vspace{-2em}
\end{table}

\section{Conclusion}

We study uncertainty estimation in VLMs for VQA, where ambiguity often arises from the visual input. Our analysis shows that decoding-based estimators can underestimate uncertainty when confident visual embeddings suppress output variation, while text perturbations and wording variations can inflate uncertainty estimates.
Motivated by these observations, we propose \textbf{Visual Semantic Entropy}, which probes uncertainty through visual perturbations and aggregates predictions at the semantic prototype level. Experiments across multiple VLMs and VQA benchmarks demonstrate that VSE provides more reliable uncertainty estimates than existing approaches.
A limitation of our method is the additional computational cost from perturbation-based sampling, which is shared with other sampling-based uncertainty estimators, as well as sensitivity to the decoding temperature used during generation.
Our approach also relies on semantic similarity models to cluster answers, and its performance may depend on the quality of the sematic distance function.
While we evaluate VSE across several representative VLMs and benchmarks, extending the analysis to larger model families and additional multimodal tasks remains an important direction for future work.
We hope this work encourages future research on uncertainty estimation that explicitly accounts for visual ambiguity.

%
%

\appendix
\title{Supplementary Material}
\author{}
\institute{}
\titlerunning{Visual Semantic Entropy}

\maketitle
\section{Implementation details}
\subsection{Algorithm}
We summarize VSE in Alg.~\ref{alg:vpd}.

\begin{algorithm}[h]
\caption{Visual Semantic Entropy (VSE)}
\label{alg:vpd}
\begin{algorithmic}[1]
\Require Question $q$, image $v$, VLM, perturbator $\mathcal{T}$, number of perturbations $M$
\Ensure Uncertainty score $\tilde{u}$

\State $a \gets \mathrm{VLM}(q, v;\, T=0)$ \Comment{Greedy decoding}

\For{$m = 1,\dots,M$}
    \State $v_m \gets \mathcal{T}(v; \xi_m)$ \Comment{Visual perturbation}
    \State $a_m \gets \mathrm{VLM}(q, v_m;T>0)$ \Comment{Sampling with pertubed image}
\EndFor

\State Cluster $\{a_m\}_{m=1}^M$ into $K$ clusters $\{c_k\}_{k=1}^K$ using $\mathrm{d}(\cdot,\cdot)$

\For{$k = 1,\dots,K$}
    \State $p_k \gets \arg\min_{a \in c_k} \sum_{a' \in c_k} \mathrm{d}(a,a')$ \Comment{Prototype assignment}
    \State $w_k \gets |c_k| / M$
\EndFor

\State $\tilde{u} \gets \sum_{k \neq k'} w_k w_{k'} \, \mathrm{d}(p_k,p_{k'})$

\State \Return $\tilde{u}$ \Comment{Higher $\tilde{u}$ indicates higher likelihood that $a$ is incorrect}
\end{algorithmic}
\end{algorithm}
\vspace{-2em}

\subsection{Datasets}
The datasets used in this work are:

\noindent\textbf{AOKVQA}: A knowledge-intensive VQA dataset where answering questions requires commonsense and world knowledge grounded in the image. It contains 17,056 / 1,145 / 6,702 samples for train/val/test; since test labels are unavailable, we evaluate on the validation split.

\noindent\textbf{MMVet}: A challenging benchmark designed to evaluate integrated multimodal reasoning abilities, combining skills such as recognition, OCR, spatial reasoning, and knowledge grounding. The benchmark contains 218 evaluation samples.

\noindent\textbf{OKVQA}: A knowledge-based VQA benchmark where answering questions requires external world knowledge beyond what is directly visible in the image, such as object functions, cultural facts, or scientific concepts. The dataset contains 14,055 questions with 9,009 / 5,046 train/test splits, and we report results on the test set.

\noindent\textbf{VILP}: A dataset probing language prior bias, where language cues suggest an answer that contradicts the visual evidence. The dataset contains 900 evaluation samples, with 300 \emph{Easy} images, each of which has 2 \emph{Adversarial} variants.

\noindent\textbf{VLM-are-biased}: A diagnostic benchmark exposing confirmation bias in VLMs through visually misleading examples that encourage reliance on memorized associations. We use the original split of 458 samples.

Together, these benchmarks evaluate uncertainty estimation across diverse VQA settings, with VILP and VLM-are-biased specifically probing visual uncertainty.

\subsection{Confident Visual Embeddings}
\begin{figure}[h]
    \centering
    \includegraphics[width=\linewidth]{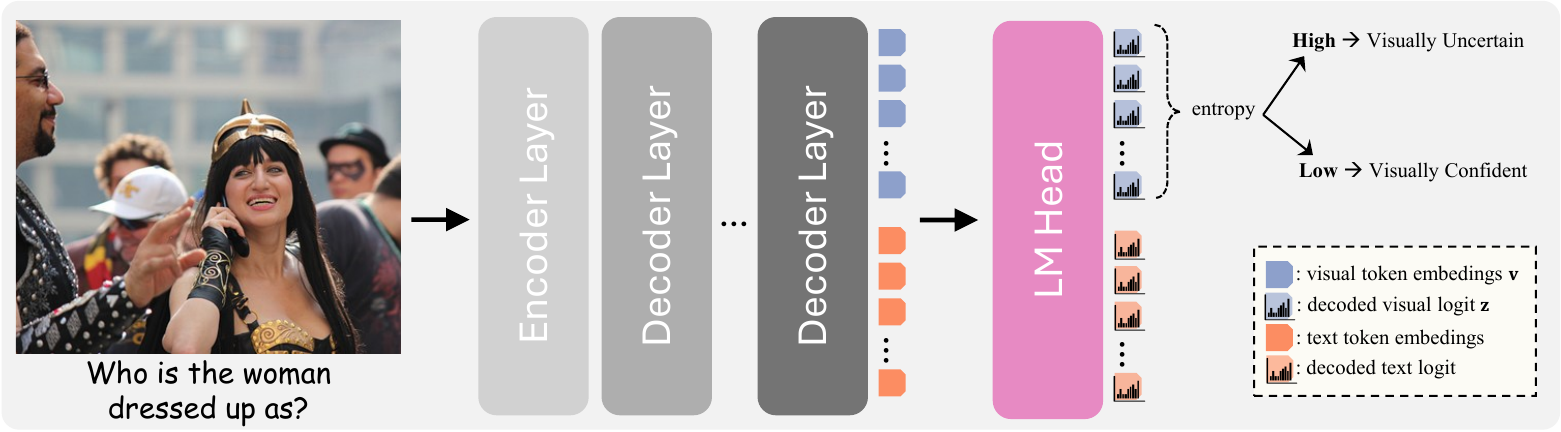}
    \caption{\textbf{Visual entropy estimation.}
Final-layer visual token embeddings are projected into the vocabulary space via the language model head (LogitLens). The entropy of the resulting token distributions is averaged to obtain the visual entropy $H_{\text{vis}}$, which measures the model's confidence in the visual input.}
    \label{fig:visual_confident_procedure}
\end{figure}

\begin{figure}
    \centering
    \includegraphics[width=\linewidth]{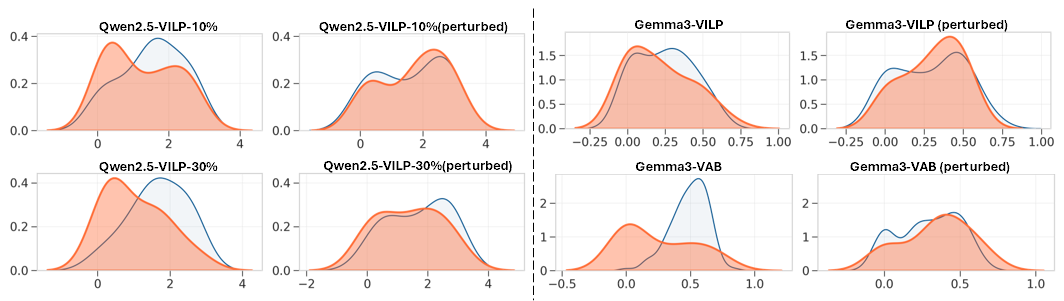}
    \caption{
\textit{Left}: Perturbation effects under 10\% \& 30\% percentiles.
\textit{Right}: Perturbation effects using Gemma3 on VILP and VAB
}
    \label{fig:vse_threshold}
\end{figure}

As illustrated in Fig.~\ref{fig:visual_confident_procedure}, we estimate the model's visual confidence by probing the final-layer visual token representations. Each visual token embedding is projected into the vocabulary space using the language model head via LogitLens~\cite{nostalgebraist2020interpreting}, producing a token distribution over the vocabulary. We compute the entropy of each projected distribution and average across visual tokens to obtain the visual entropy $H_{\text{vis}}$, which reflects the model's confidence in the visual representation. This metric allows us to distinguish visually confident inputs from visually uncertain ones, enabling us to analyze cases where semantic entropy fails to reflect uncertainty despite ambiguous visual evidence.
We use the top/bottom 20\% percentile to define high and low visual entropy. 
We additionally test top/bot. 10\% \& 30\% percentiles and also observe increased entropy under perturbations (Fig.~\ref{fig:vse_threshold}-\textit{Left}).

\subsubsection{Robustness.}
We further analyze the observation in Sec.~3.2 using Qwen2.5-VL and Gemma3 on VILP and VAB (Fig.~\ref{fig:vse_threshold}-\textit{Right}).
Quantitatively, we measure the entropy increase
and report the mean $\Delta H$ with 1-sided Wilcoxon signed-rank test $p$-value in Table.~\ref{tab:delta_h}.
We consistently observe positive entropy shifts, with most being statistically significant, suggesting robustness across datasets and models. 

\begin{table}[t]
\centering
\caption{$\Delta H$ under perturbations.}
\label{tab:delta_h}
\resizebox{0.95\linewidth}{!}{
\begin{tabular}{c|cccc}
& \textbf{VILP-Qwen} 
& \textbf{VILP-Gemma} 
& \textbf{VAB-Qwen} 
& \textbf{VAB-Gemma} \\
\midrule
\textbf{$\Delta H$ ($p$)} 
& 0.1625 (\textbf{0.0203}) & 0.0939 (\textbf{0.0238}) & 0.0750 (\textbf{0.0360}) & 0.0923 (0.1057) \\
\end{tabular}
}
\end{table}

\section{Visual Ambiguity analysis.}

We understand that visual ambiguity is semantically related, and therefore analyze whether low-level VSE relates to semantic regions in pixel space (feature-level analysis is included in Sec.~\ref{sec:feature_perturb}).
We identify \emph{semantic regions} using SAM3 to localize objects mentioned in the questions and answers in AOKVQA, then apply Gaussian noise either in/outside these regions.
The AUC of 
\textit{Inside},
\textit{Outside} and 
\textit{Full}-image 
VSE are
\textbf{79.1},\textbf{75.2},\textbf{78.3}
respectively.
Moreover, full-image VSE strongly correlates with semantic-region VSE ($r$=0.6), suggesting that full-image VSE are driven by semantic regions.
Therefore, although Gaussian noise is low-level, VSE reflects visual ambiguity rather than noise-induced instability.

\section{Locality of Text and Images}
\begin{figure}[h]
    \centering
    \begin{subfigure}{0.32\textwidth}
        \centering
        \includegraphics[width=\linewidth]{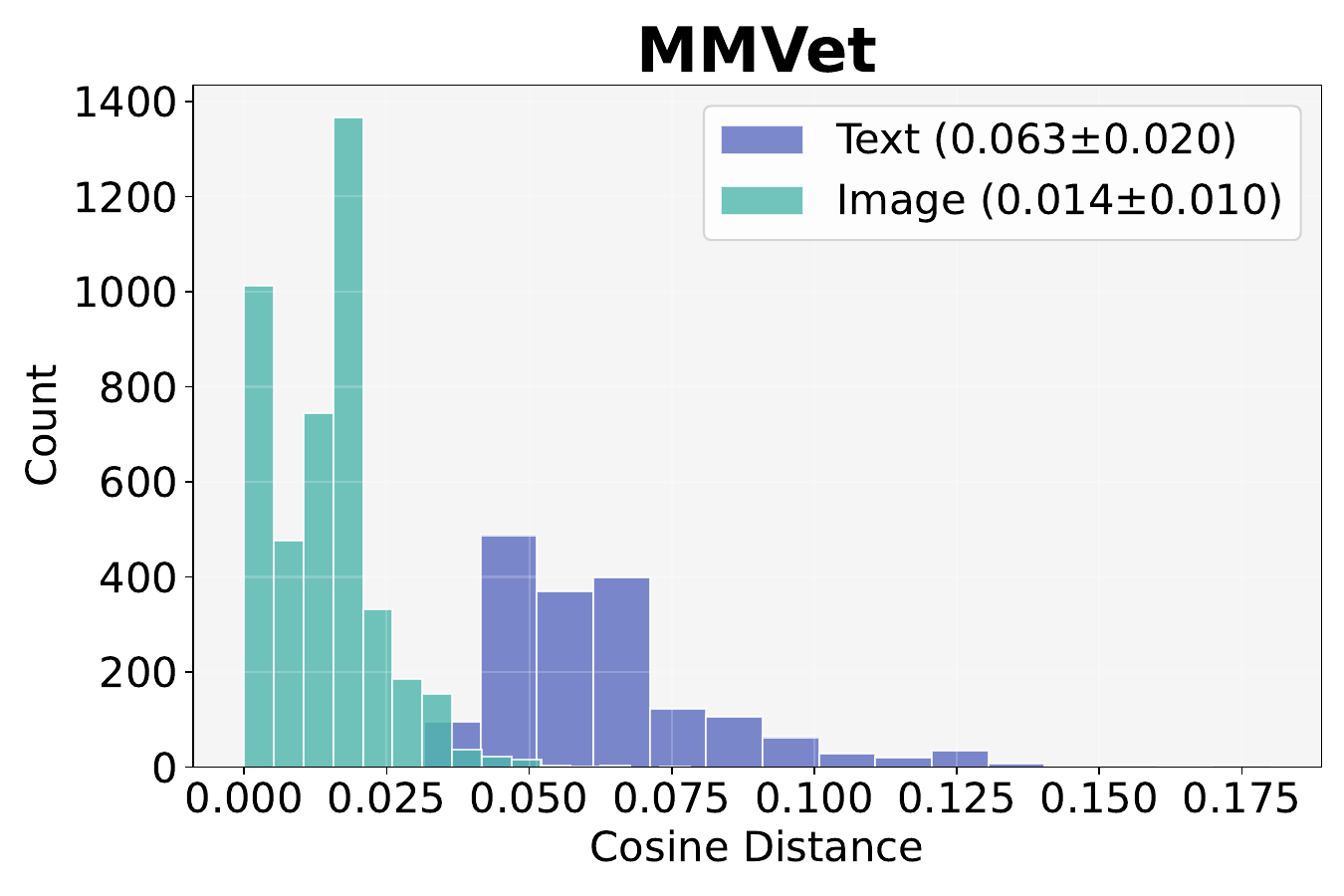}
    \end{subfigure}
    \hfill
    \begin{subfigure}{0.32\textwidth}
        \centering
        \includegraphics[width=\linewidth]{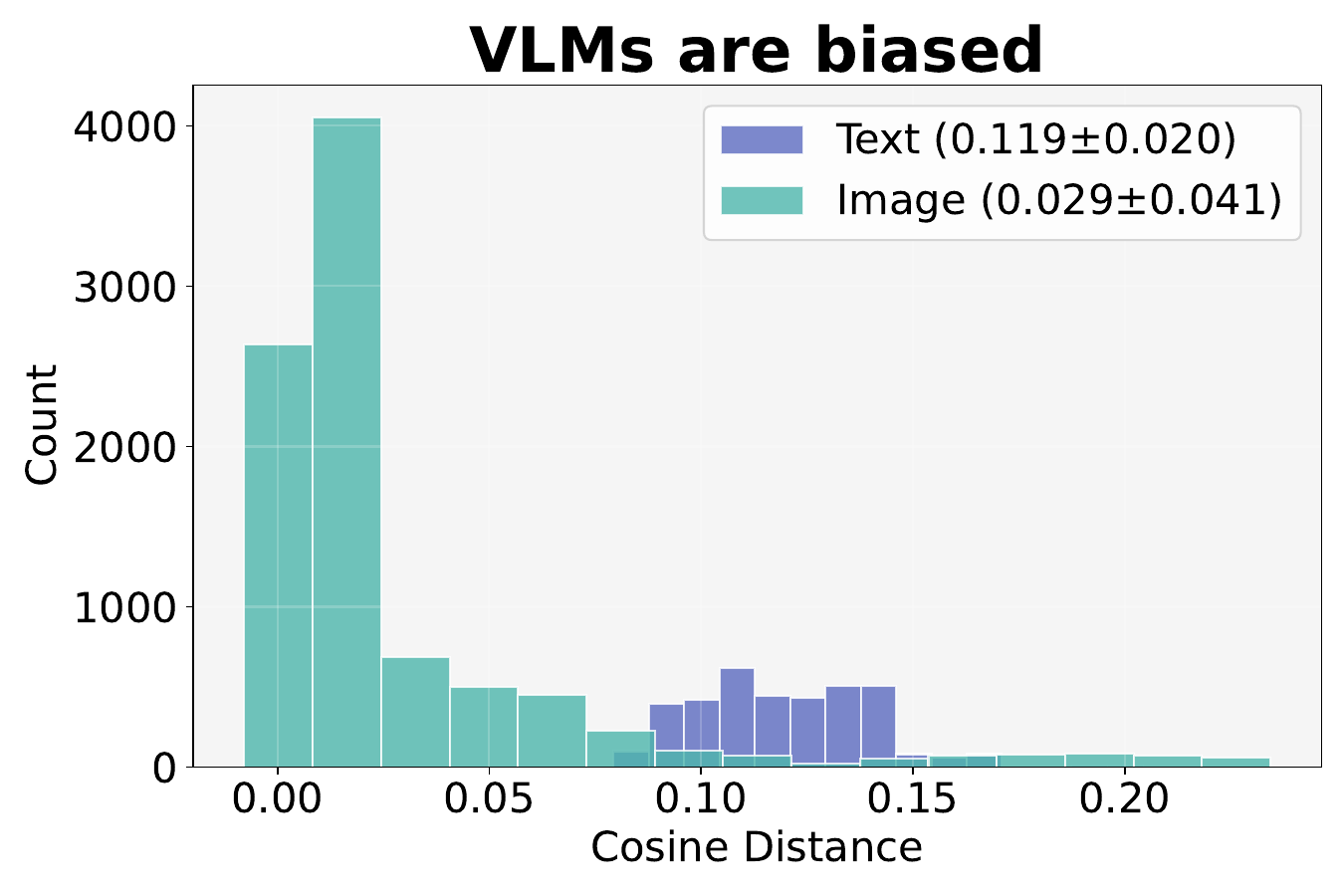}
    \end{subfigure}
    \hfill
    \begin{subfigure}{0.32\textwidth}
        \centering
        \includegraphics[width=\linewidth]{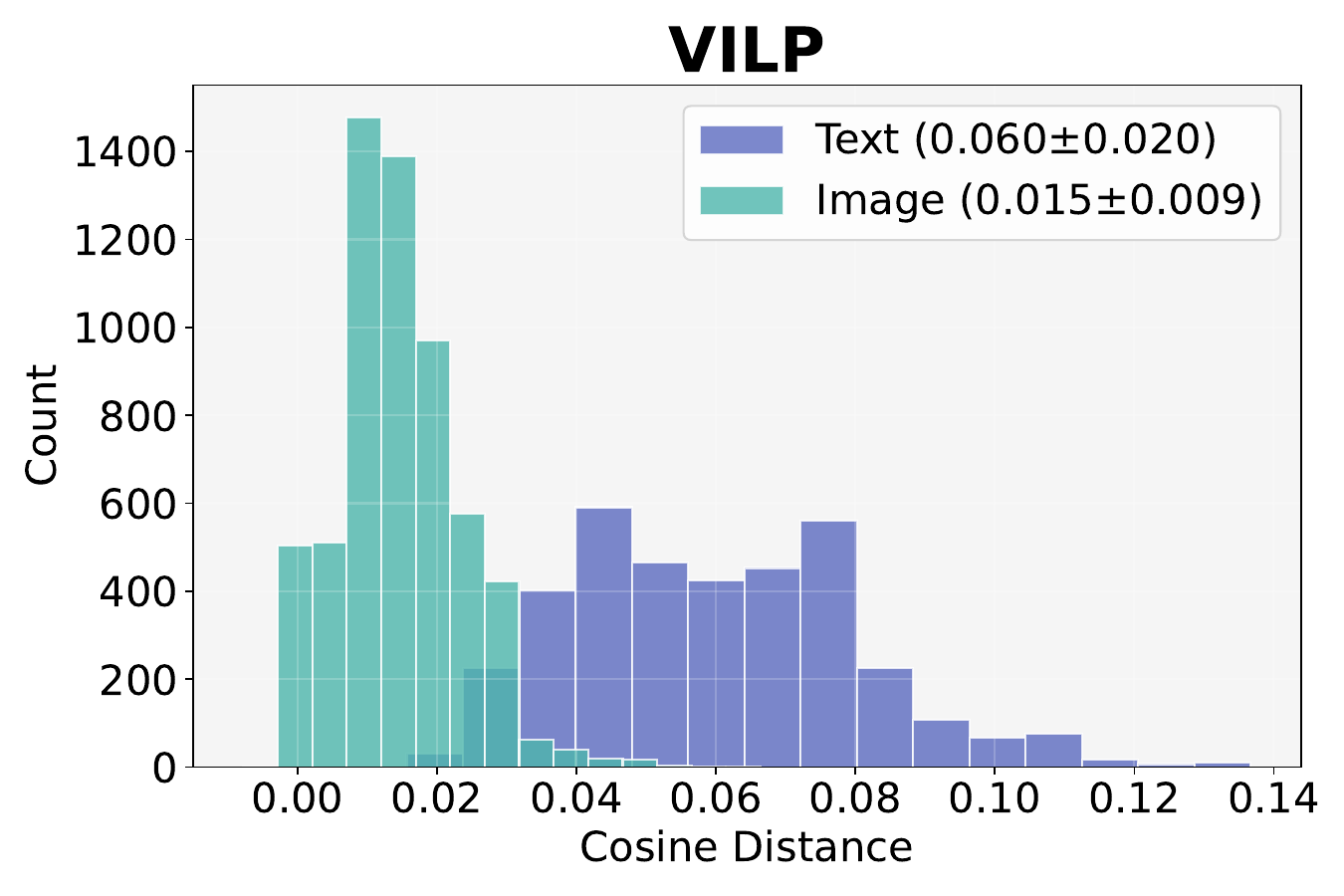}
    \end{subfigure}

    \caption{\textbf{Embedding shifts induced by text and image perturbations.}
For each image–question pair, we generate semantically equivalent paraphrases and image perturbations, then encode all inputs using the Qwen3-VL-Embedding-8B multimodal embedding model. The cosine distance between each perturbed embedding and the original embedding is measured. Across datasets, textual paraphrases produce substantially larger embedding shifts than image perturbations: $0.063$ vs.\ $0.014$ on MMVet, $0.119$ vs.\ $0.029$ on VLMs Are Biased, and $0.060$ vs.\ $0.015$ on VILP.}
    \label{fig:distance_dist}
    \vspace{-1em}
\end{figure}

\noindent\textbf{Image vs.\ Text Embedding shifts.} To examine how different perturbations affect the multimodal representation, we compare the embedding shifts induced by textual paraphrases and image perturbations. For each question, we generate semantically equivalent paraphrases using Gemma3-8B~\cite{gemma2025gemma3} following the prompting strategy of~\cite{zhang2024vl}. Image perturbations are produced by adding Gaussian noise with standard deviation 20 to the input image. All perturbed inputs are then encoded into a shared multimodal embedding space using Qwen3-VL-Embedding-8B~\cite{li2026qwen3}.
For each original image–question pair, we generate eight image perturbations and eight question paraphrases, and compute the cosine distance between their embeddings and the embedding of the original pair.
Fig.~\ref{fig:distance_dist} shows the resulting distance distributions across datasets. Image perturbations consistently induce small embedding shifts, indicating that they explore a local neighborhood around the original query. In contrast, textual paraphrases produce substantially larger shifts, suggesting that paraphrasing moves the representation to distant regions of the embedding space rather than sampling locally around the original conditioning.

\begin{figure}[h]
    \centering
    
    \begin{subfigure}{0.32\textwidth}
        \centering
        \includegraphics[width=\linewidth]{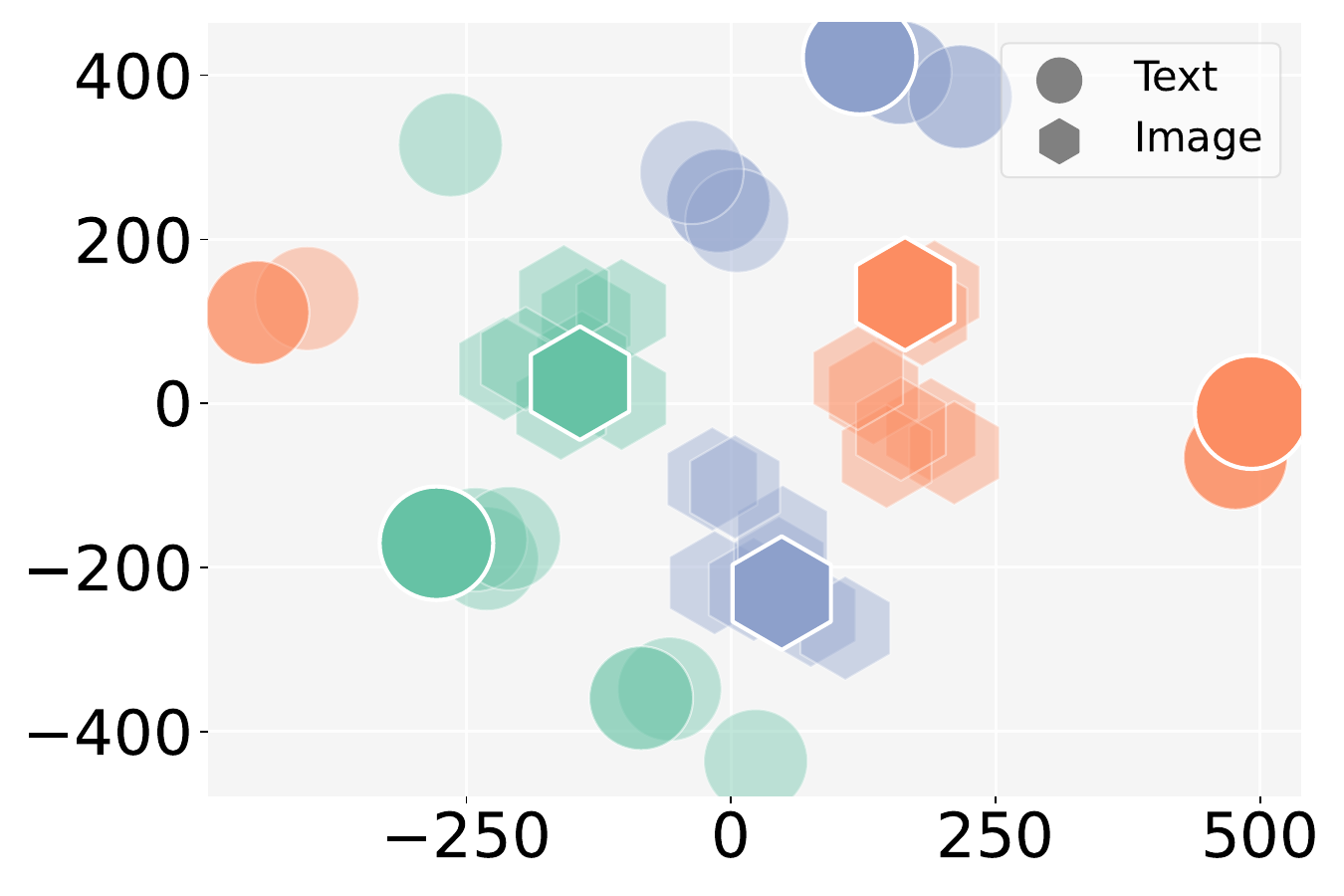}
    \end{subfigure}
    \hfill
    \begin{subfigure}{0.32\textwidth}
        \centering
        \includegraphics[width=\linewidth]{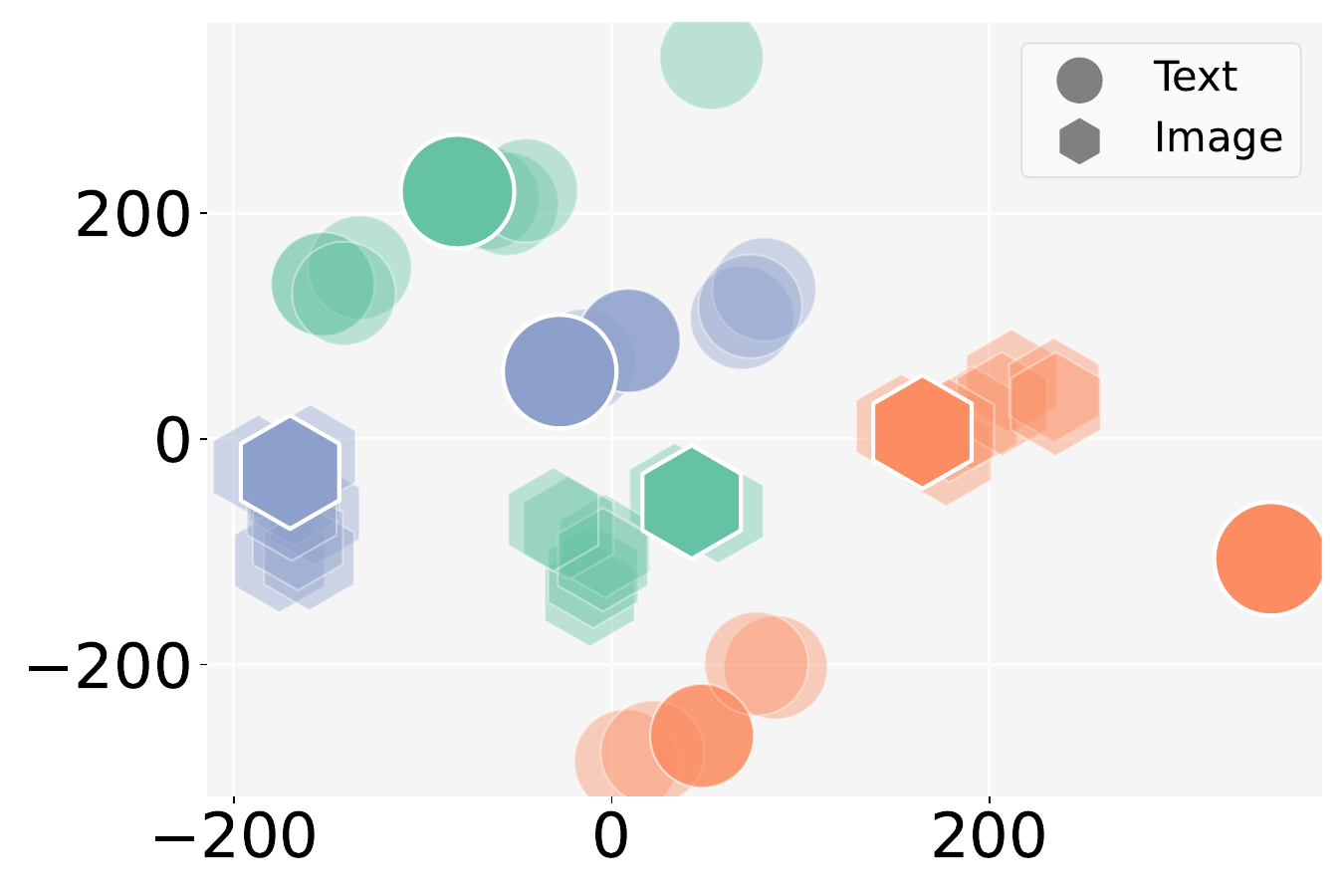}
    \end{subfigure}
    \hfill
    \begin{subfigure}{0.32\textwidth}
        \centering
        \includegraphics[width=\linewidth]{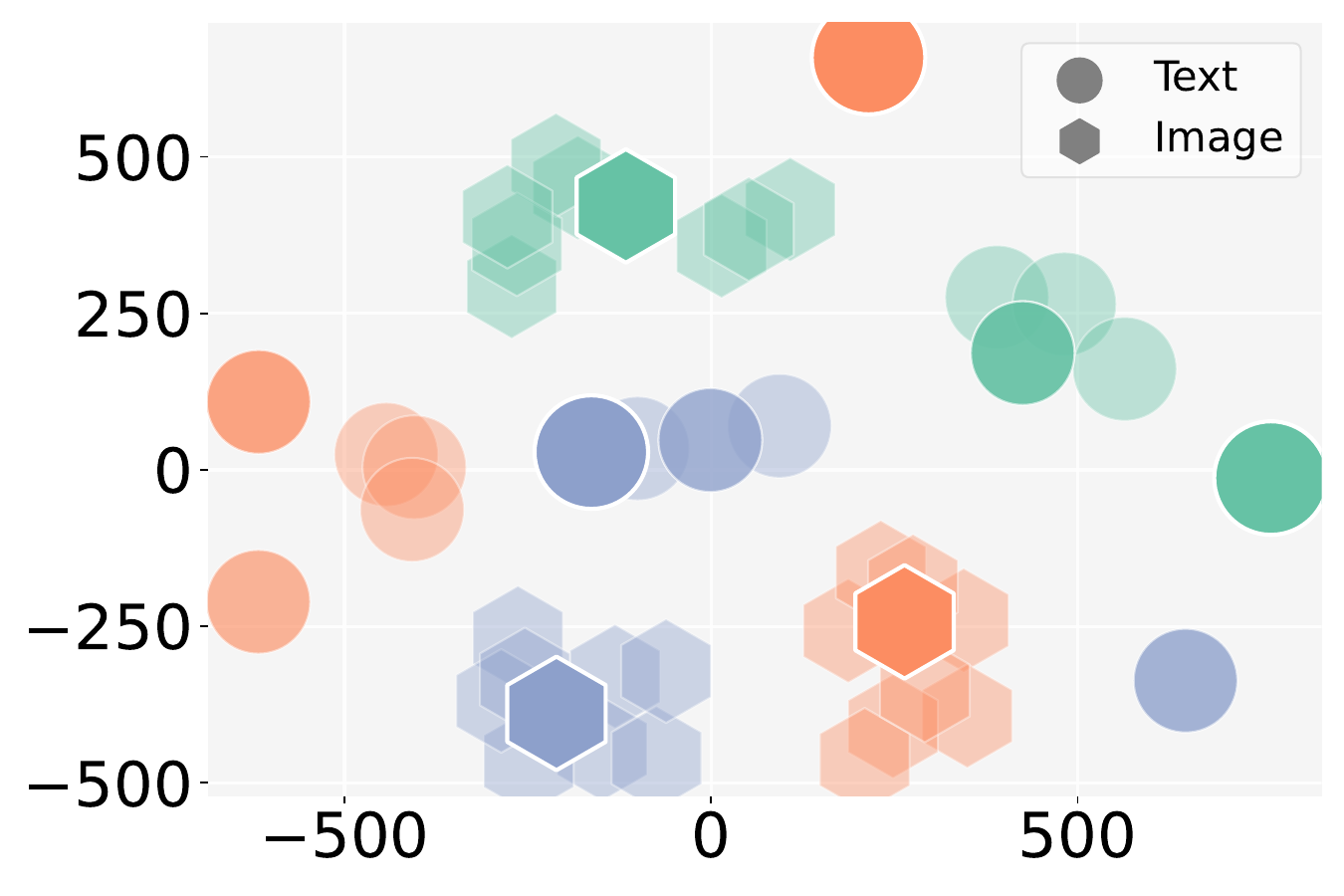}
    \end{subfigure}

    \caption{\textbf{Image vs.\ Text Embedding Space.}
t-SNE projections of multimodal embeddings for several image–question pairs from AOKVQA dataset. Colors indicate the same original image-question pair, while marker shapes distinguish perturbation types (text vs.\ image). Image perturbations remain tightly clustered around the original embedding, whereas question paraphrases induce larger shifts and spread to distant regions of the embedding space.}
    \label{fig:multimodal_space}
\end{figure}

\noindent\textbf{Image vs.\ Text Embedding Space.} We further visualize the multimodal embedding space using t-SNE. As shown in Fig.~\ref{fig:multimodal_space}, image perturbations remain close to the original embedding, whereas question paraphrases induce substantially larger shifts and disperse across the space.

\section{Text perturbation drives cluster assignments}

\begin{figure}[h]
    \centering
    
    \begin{subfigure}{0.48\textwidth}
        \centering
        \includegraphics[width=\linewidth]{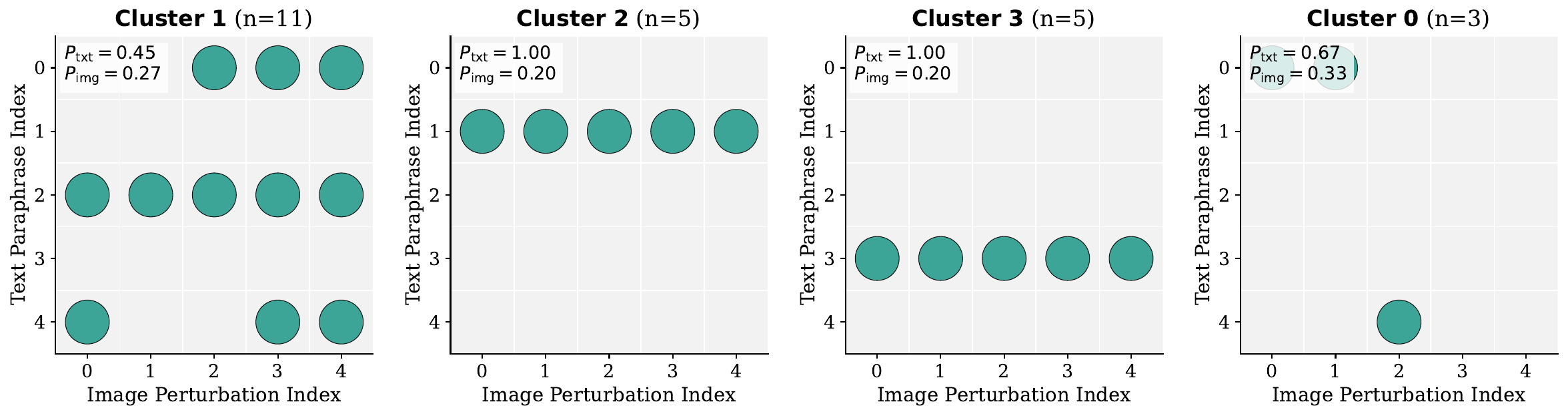}
    \end{subfigure}
    \hfill
    \begin{subfigure}{0.48\textwidth}
        \centering
        \includegraphics[width=\linewidth]{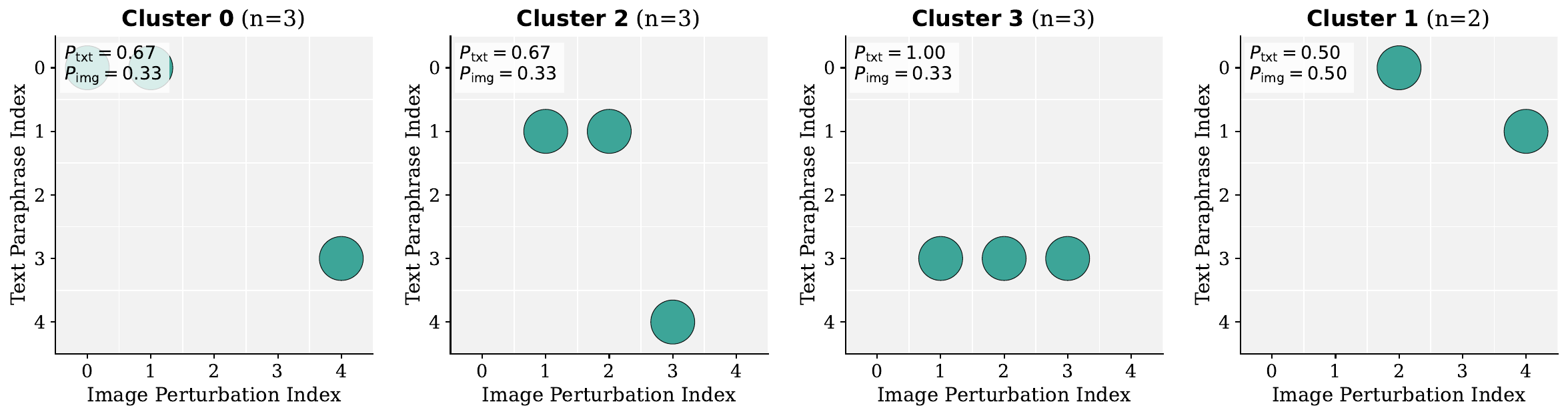}
    \end{subfigure}
    
    \begin{subfigure}{0.48\textwidth}
        \centering
        \includegraphics[width=\linewidth]{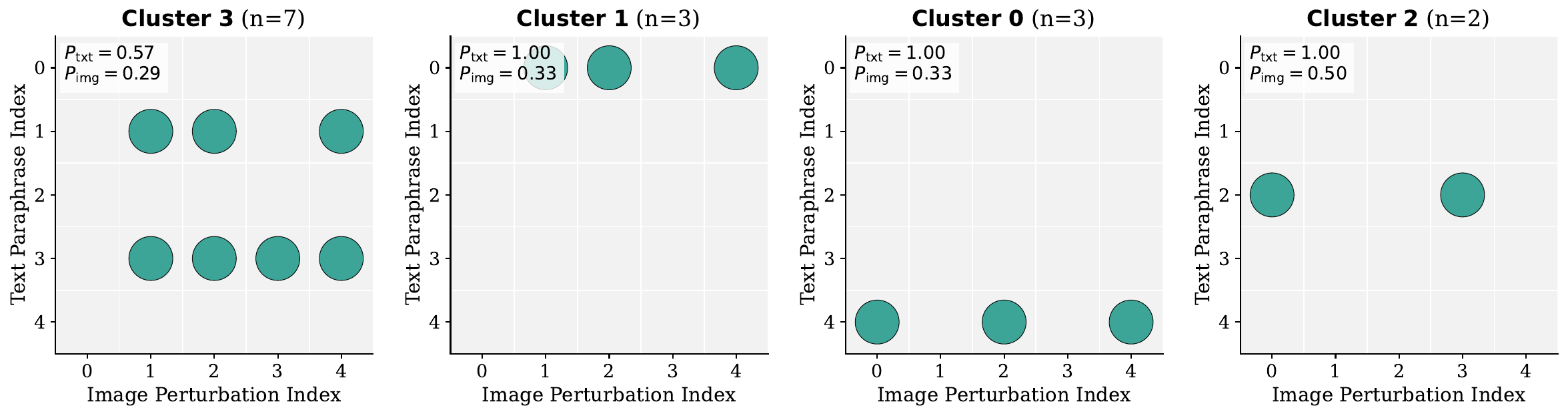}
    \end{subfigure}
    \hfill
    \begin{subfigure}{0.48\textwidth}
        \centering
        \includegraphics[width=\linewidth]{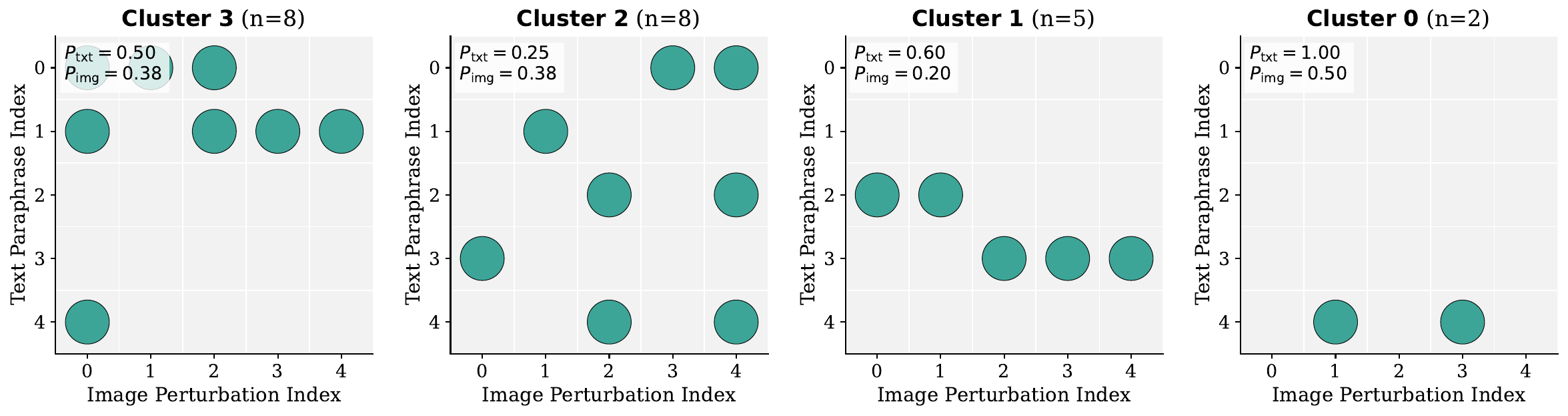}
    \end{subfigure}

    \begin{subfigure}{0.48\textwidth}
        \centering
        \includegraphics[width=\linewidth]{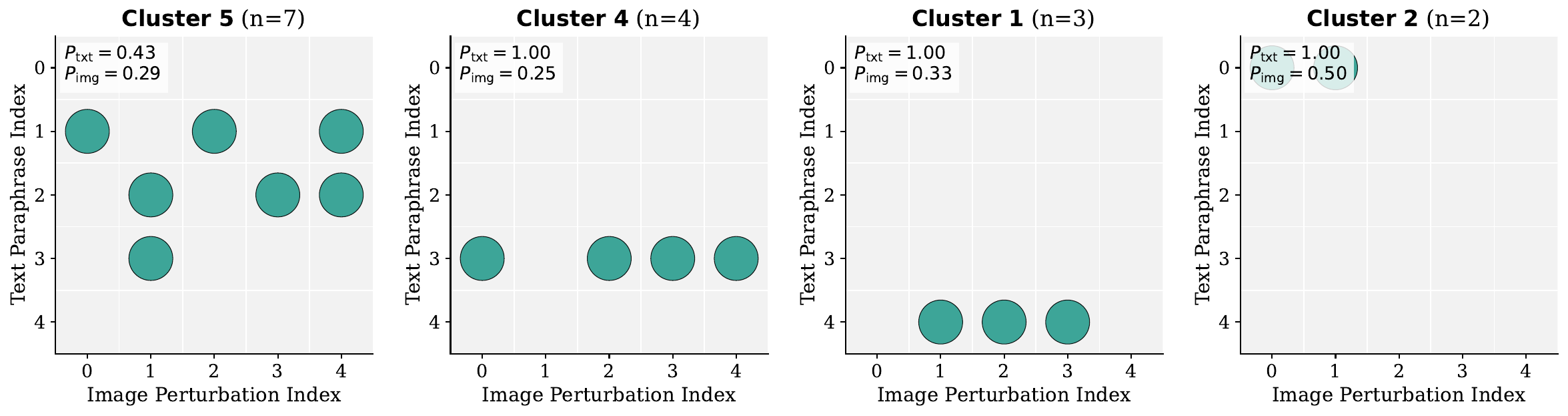}
    \end{subfigure}
    \hfill
    \begin{subfigure}{0.48\textwidth}
        \centering
        \includegraphics[width=\linewidth]{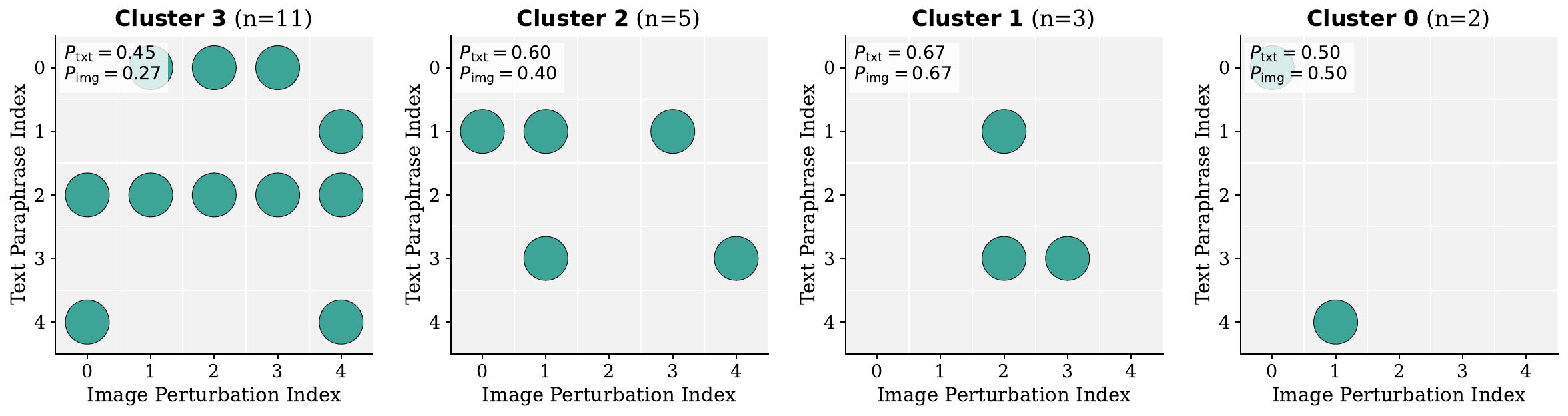}
    \end{subfigure}
    
    \begin{subfigure}{0.48\textwidth}
        \centering
        \includegraphics[width=\linewidth]{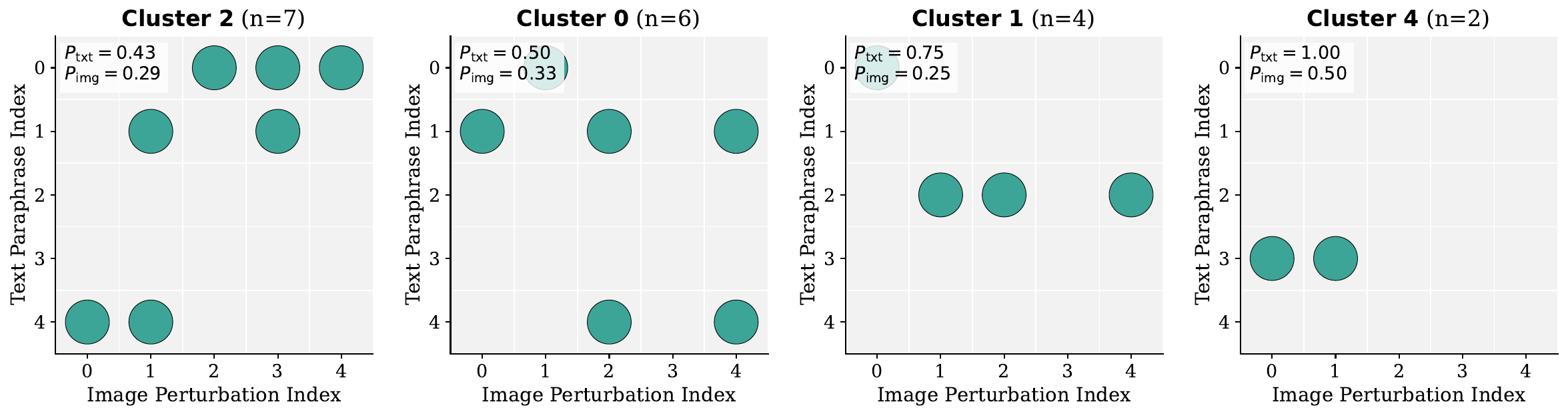}
    \end{subfigure}
    \hfill
    \begin{subfigure}{0.48\textwidth}
        \centering
        \includegraphics[width=\linewidth]{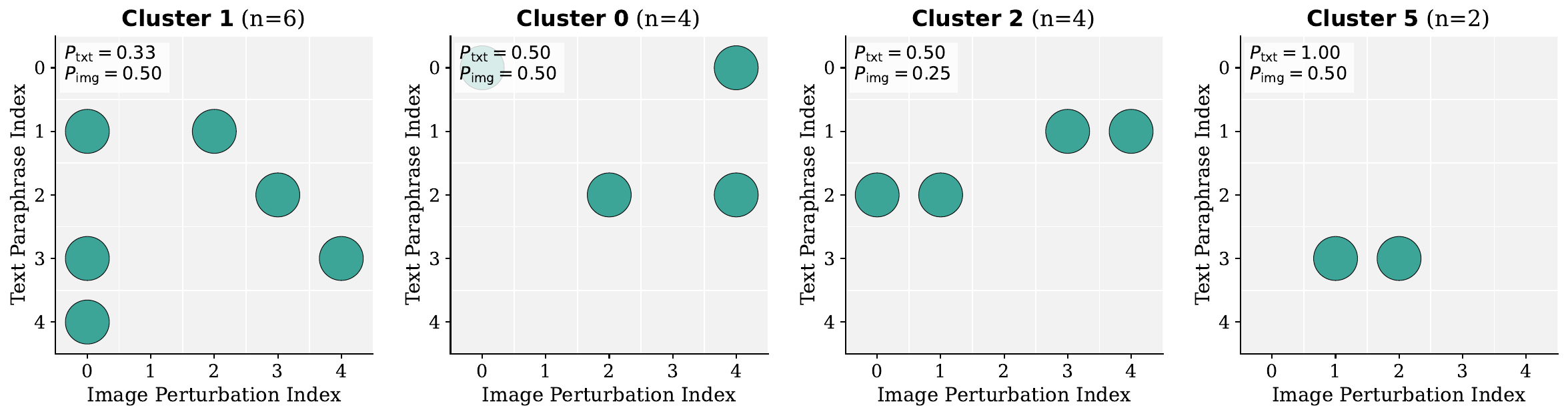}
    \end{subfigure}

    \begin{subfigure}{0.48\textwidth}
        \centering
        \includegraphics[width=\linewidth]{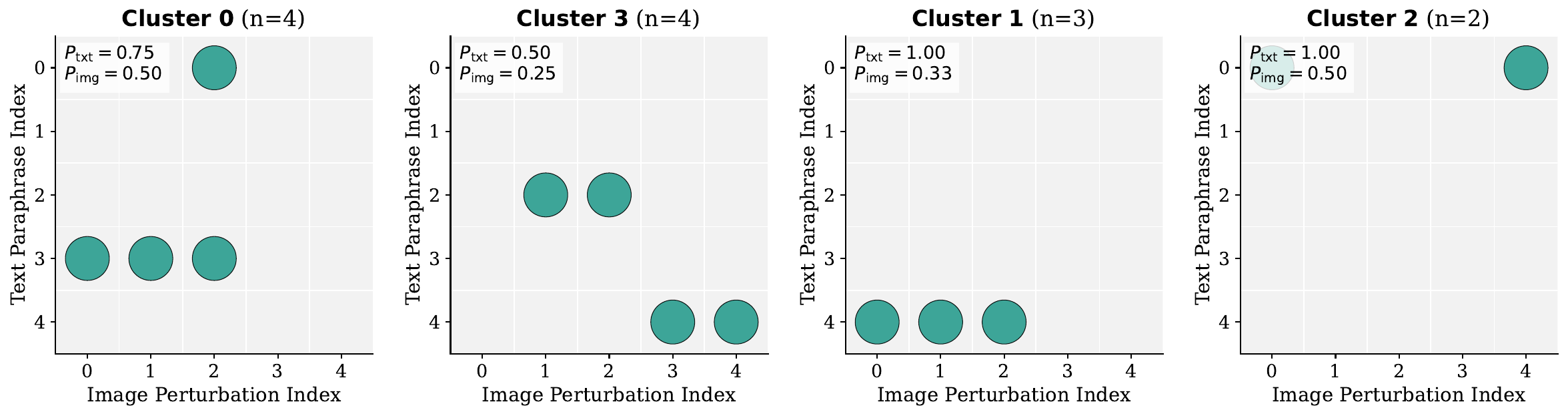}
    \end{subfigure}
    \hfill
    \begin{subfigure}{0.48\textwidth}
        \centering
        \includegraphics[width=\linewidth]{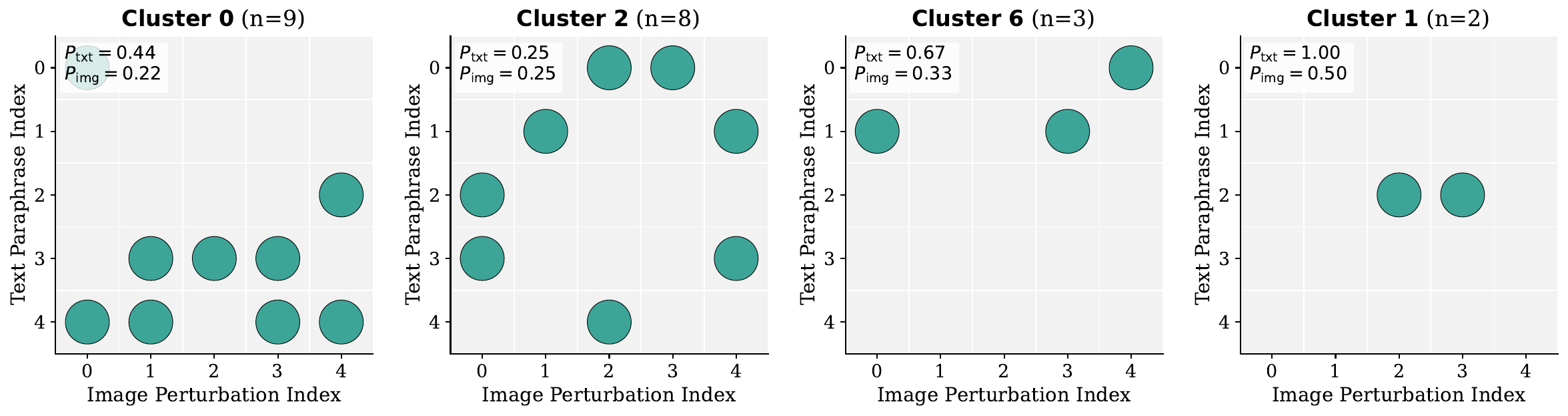}
    \end{subfigure}
    \caption{\textbf{Cluster occupancy map.}
Each panel shows cluster assignments on the $L \times M$ perturbation grid  of a random sample (rows: textual paraphrases; columns: image perturbations). Horizontal stripes indicate invariance across image perturbations, showing that clustering is dominant by textual paraphrase. (AOKVQA dataset, Qwen2.5-VL).}
\label{fig:cluster_map_2}
\end{figure}

Such non-local shifts have important implications for uncertainty estimation. When both image and text are perturbed, outputs produced under different paraphrases are likely to occupy distinct regions of the embedding space, even if the visual information remains unchanged. As a result, clustering would group outputs by paraphrase identity rather than by variations in the visual input.
To verify this effect, we further provide qualitative examples of the cluster occupancy patterns on the perturbation grid (Fig.~\ref{fig:cluster_map_2}). We observe horizontal stripe structures where cluster membership is consistent within a paraphrase but invariant across image perturbations. This pattern confirms that cluster assignments are largely driven by the textual input.
Hence, these results show that joint image-text perturbation primarily introduces variability through textual paraphrases rather than visual changes. Consequently, uncertainty estimates derived from such perturbations may reflect prompt-induced variability instead of genuine visual ambiguity.

\section{Hyperparameters Sensitivity}

\begin{figure}[h]
    \centering
    \begin{subfigure}[b]{0.49\textwidth}
        \centering
        \includegraphics[width=\textwidth]{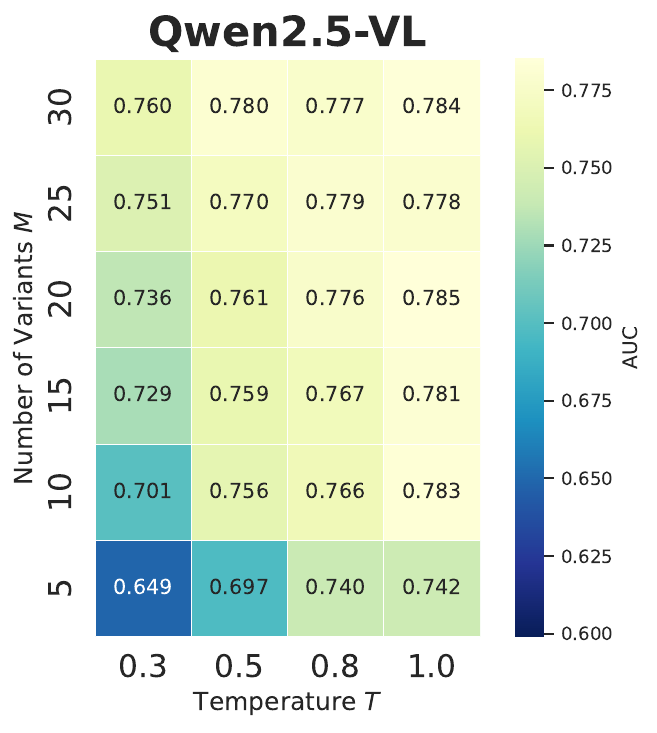}
    \end{subfigure}
    \hfill
    \begin{subfigure}[b]{0.49\textwidth}
        \centering
        \includegraphics[width=\textwidth]{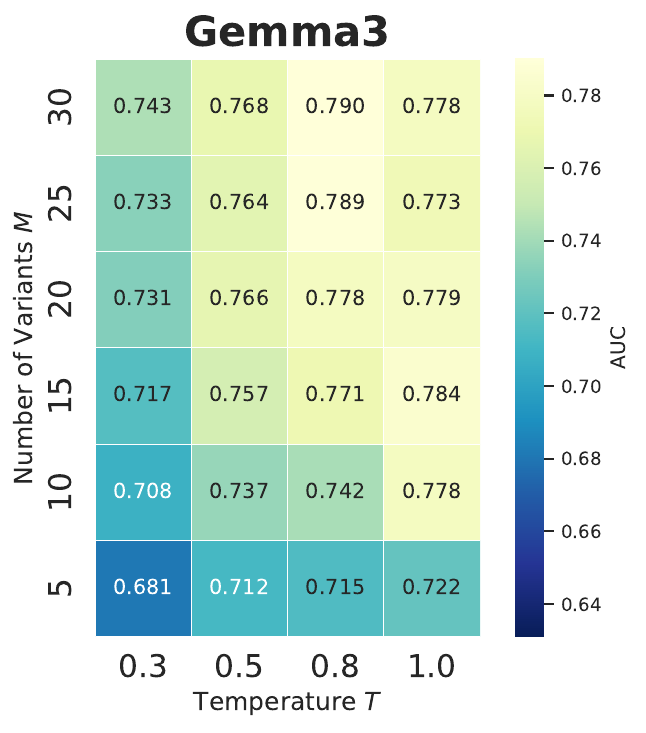}
    \end{subfigure}

    \vspace{-1em}
    \caption{\textbf{Decoding temperature $T$ and Number of variants $M$ ablation.} Performance improves with more variants and moderate temperatures. Results on AOKVQA dataset.}
    \label{fig:ablation_heatmap}
    \vspace{-2em}
\end{figure}

\subsection{Temperature $T$ and Number of samples $M$}

\begin{figure}[h]
    \centering
    \begin{subfigure}[b]{0.4\textwidth}
        \centering
        \includegraphics[width=\textwidth]{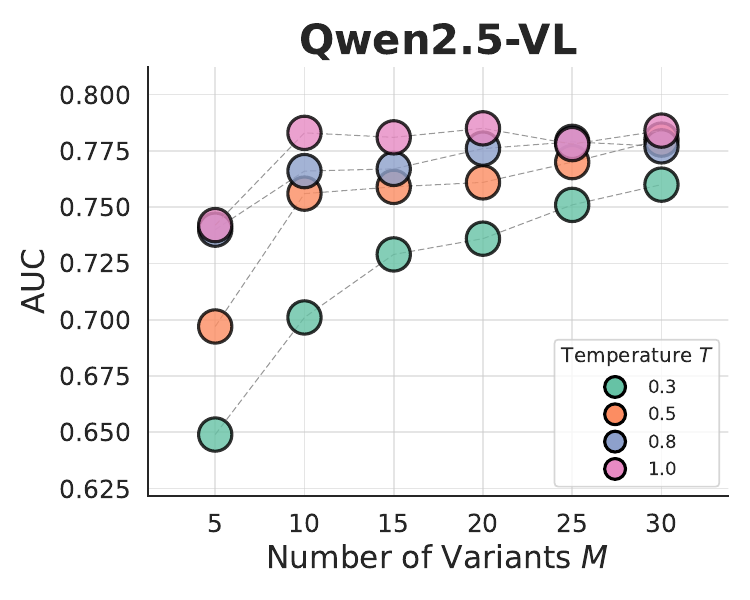}
    \end{subfigure}
    \begin{subfigure}[b]{0.4\textwidth}
        \centering
        \includegraphics[width=\textwidth]{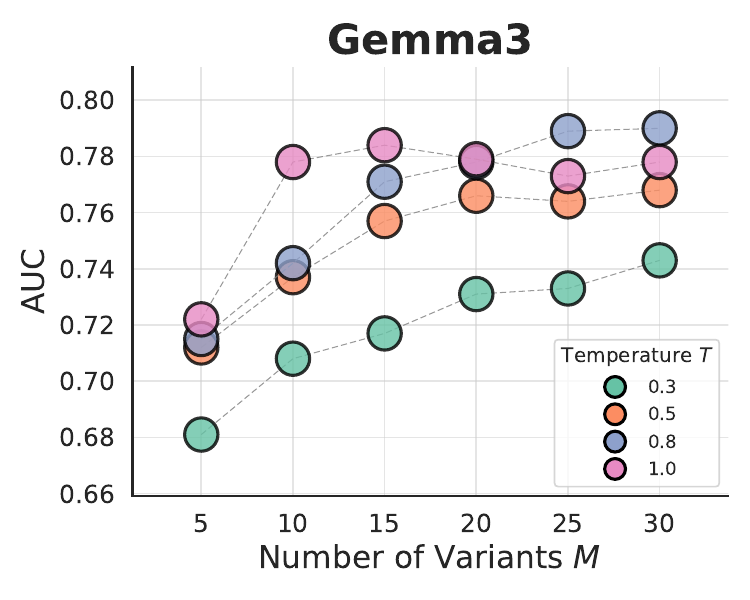}
    \end{subfigure}
    \vspace{-1em}
    \caption{Adequate sampling ($M$) and high temperature ($T$)  demonstrate improved performance, with gains saturating as $M$ increases. Results on AOKVQA dataset.}
    \label{fig:ablation_line}
\end{figure}

Fig.~\ref{fig:ablation_heatmap} analyzes the impact of decoding temperature $T$ and the number of generated variants $M$ on uncertainty estimation performance. Increasing $M$ consistently improves AUC for both Qwen2.5-VL and Gemma3, with the largest gains observed from 5 to around 15-20 samples. Performance largely saturates once $M\geq 15$, despite the linear increase in computational cost, indicating diminishing returns from additional sampling.
Higher temperatures ($T \in [0.8, 1.0]$) consistently achieve the best results, while lower temperatures suppress output diversity and limit uncertainty signals. Overall, effective uncertainty estimation benefits from sufficiently stochastic decoding and a moderate number of samples, balancing diversity and computational efficiency.
Fig.~\ref{fig:ablation_line} presents the same ablation in line-plot form for clarity. Increasing the number of variants $M$ and decoding temperature $T$ consistently improves AUC, with gains diminishing beyond $M \geq 15$.

\subsection{Augmentation strength $\xi$}

\begin{figure}[h]
    \centering
    \begin{subfigure}[b]{0.4\textwidth}
        \centering
        \includegraphics[width=\textwidth]{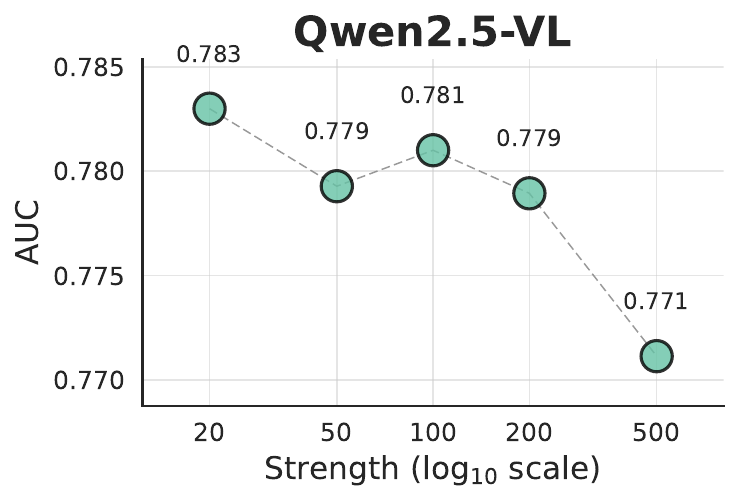}
    \end{subfigure}
    \begin{subfigure}[b]{0.4\textwidth}
        \centering
        \includegraphics[width=\textwidth]{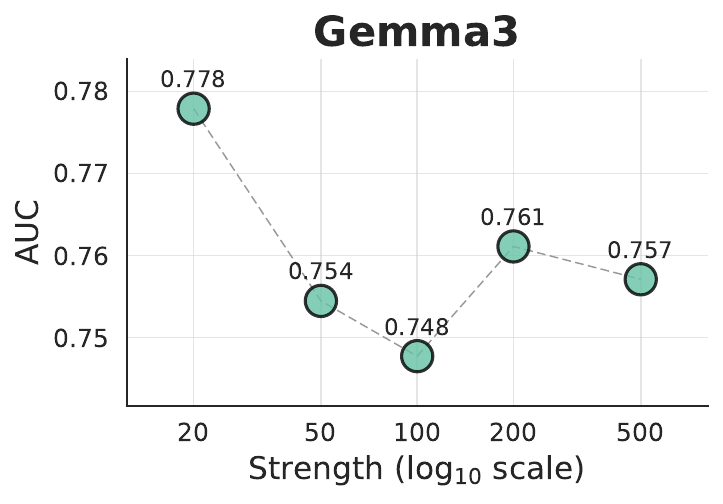}
    \end{subfigure}

    \caption{\textbf{Visual perturbation strength $\xi$ ablation.} Moderate noise improves performance, while stronger perturbations cause slight degradation. Results on AOKVQA dataset.}
    \label{fig:ablation_strength}
\end{figure}

We vary the visual perturbation strength from 20 to 500 and observe that moderate noise at 20 yields the best performance, while stronger perturbations slightly reduce AUC (Fig.~\ref{fig:ablation_strength}). This falls into the behavior of text-driven perturbations: large textual shifts introduce variability that is not grounded in visual evidence and are therefore harmful. Similarly, overly strong visual perturbations create excessive shifts in the visual space, distorting semantic content rather than revealing genuine ambiguity. In both cases, large shifts degrade uncertainty estimation quality.

\begin{table}[h]
\centering
\caption{\textbf{Visual Perturbation Functions $\mathcal{T}$ ablation}. Results on AOKVQA dataset.}
\label{tab:augmentation_ablation}
\begin{tabular}{lcccc}
\toprule
Model & \shortstack{Affine\\Transform} & \shortstack{Color\\Transform} & \shortstack{Gaussian\\Blur} & \shortstack{Gaussian\\Noise} \\
\midrule
\textbf{Qwen2.5-VL} & 0.759 & \underline{0.764} & 0.749 & \textbf{0.783} \\
\textbf{Gemma3}     & 0.742 & 0.754 & \textbf{0.782} & \underline{0.778} \\
\bottomrule
\end{tabular}
\end{table}

\subsection{Visual Perturbation Functions $\mathcal{T}$}

Table~\ref{tab:augmentation_ablation} compares different visual perturbation functions for uncertainty estimatio (AUC) on AOKVQA. We evaluate four augmentation types with controlled magnitudes:

\begin{itemize}
    \item \textbf{Affine Transform:} small geometric perturbations including rotation ($\pm 10^\circ$), translation (up to $5\%$ of image size), scaling within $[0.95, 1.05]$ ratio, and shear ($\pm 10^\circ$).
    
    \item \textbf{Color Transform:} photometric adjustments implemented via random brightness, contrast, and saturation scaling sampled from $[0.6, 1.4]$, and hue shift sampled from $[-0.02, 0.02]$ in normalized HSV space.
    \item \textbf{Gaussian Blur:} spatial smoothing using a Gaussian kernel with a small blur radius ($1\%$ image size) and mild spatial jitter ($0.1$).
    \item \textbf{Gaussian Noise:} additive pixel-wise Gaussian noise with standard deviation $\sigma = 20$, applied independently to each pixel.

\end{itemize}

Across both Qwen2.5-VL and Gemma3, Gaussian-based perturbations (noise and blur) achieve stronger AUC compared to affine and color transformations. While the differences are moderate, Gaussian noise yields the best result for Qwen2.5-VL and Gaussian blur performs best for Gemma3. 
Gaussian-based perturbations (noise and blur) consistently achieve stronger performance across both models. This likely stems from their ability to introduce local, distributed variations that preserve the global semantic structure of the image while probing its immediate neighborhood in representation space. We adopt Gaussian noise as the default perturbation function, as it consistently performs well across models and requires tuning only a single parameter, the noise standard deviation $\sigma$. 

\subsubsection{Feature Perturbation}
\label{sec:feature_perturb}
We also implemented masked patches and feature-level visual-token perturbations, which perform comparably or slightly higher than Gaussian noise. Results on AOKVQA:

\begin{table}[h]
\centering
\caption{\textbf{Feature Perturbation ablation}. Results on AOKVQA dataset.}
\label{tab:feature_perturbation_ablation}
\begin{tabular}{lccc}
\toprule
Model & \shortstack{Gaussian\\Noise} & \shortstack{Feature\\Perturbation} & \shortstack{Patch\\Mask} \\
\midrule
\textbf{Qwen2.5-VL} & 0.783 & \underline{0.785} & \textbf{0.793} \\
\textbf{Gemma3}     & 0.778 & \underline{0.780} & \textbf{0.783} \\
\bottomrule
\end{tabular}
\end{table}

\noindent As magnitudes are not directly comparable, we ran each method over a strength sweep (mask ratio [10\%, 40\%], noise scale [2\%, 10\%]) and report best result.
We observe that: 

\textbf{(1)} VSE remains strong across perturbation types and not tied to low-level perturbation

\textbf{(2)} Gaussian noise achieves competitive performance, being a simple and practical probe.

We note that all compared methods are blackbox without feature access, and VSE is designed under the same constraint. We agree that feature-level analysis is important and will include it in the paper.

\subsection{Distance Function d(,)}

\begin{table}[h]
\centering
\caption{\textbf{Semantic Distance Function d(,) ablation.} AUC comparison between SNNE and VSE under different semantic distance function $\text{d}(,)$. Results on AOKVQA dataset.}
\label{tab:snne_vpd}
\begin{tabular}{lcccccc}
\toprule
& \multicolumn{3}{c}{\textbf{SNNE}} & \multicolumn{3}{c}{\textbf{VSE}} \\
\cmidrule(lr){2-4} \cmidrule(lr){5-7}
Model & \small{Cosine} & \small{BertScore} & \small{DeBERTa} & \small{Cosine} & \small{BertScore} & \small{DeBERTa} \\
\midrule
\textbf{Qwen2.5-VL} & 0.651 & \underline{0.742} & \textbf{0.744} & 0.685 & \underline{0.754} & \textbf{0.783} \\
\textbf{Gemma3}     & 0.614 & \textbf{0.715} & \underline{0.700} & 0.657 & \underline{0.720} & \textbf{0.778} \\
\bottomrule
\end{tabular}
\vspace{-2em}
\end{table}

Table~\ref{tab:snne_vpd} compares SNNE and VSE under different semantic similarity metrics, including:
\begin{itemize}
    \item Cosine distance with all-MiniLM-L6-v2 embedding model~\cite{wang2020minilm}
    \item BERTScore~\cite{zhang2019bertscore},
    \item DeBERTa-v2-xlarge-mnl~\cite{he2020deberta}.
\end{itemize}

Across both Qwen2.5-VL and Gemma3, VSE consistently outperforms SNNE under stronger semantic metrics. In particular, when using DeBERTa, VSE achieves the highest AUC for both models (0.783 for Qwen2.5-VL and 0.778 for Gemma3).
While cosine similarity provides weaker alignment signals, performance improves when adopting contextualized semantic metrics such as\\ BERTScore and DeBERTa. The gains are more pronounced for VSE, suggesting that clustering-based uncertainty estimation benefits from richer semantic representations. Overall, these results indicate that combining VSE with a strong semantic similarity measure yields the most reliable uncertainty estimates.

\section{Complexity}
Complexity
of VSE, SE, and SNNE is 
$O(M^2)$
. Inference is dominated by VLM sampling, which are all similar, while the aggregation step is negligible:
0.24/0.34/0.21s.

\section{Qualitative Examples}
\begin{figure}
    \centering
    \includegraphics[width=\linewidth]{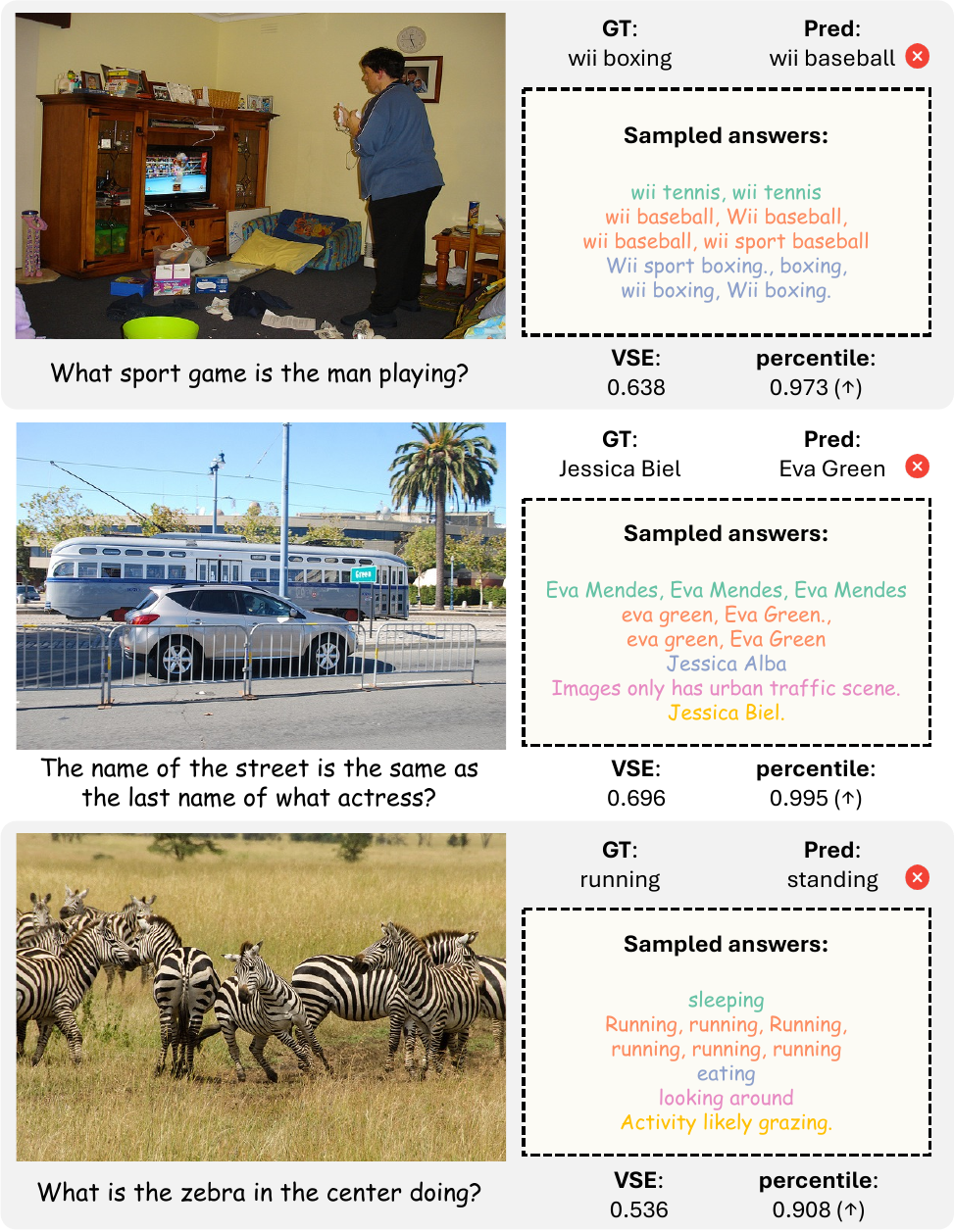}
    \caption{\textbf{Qualitative Results.} A high VSE indicates that the model is uncertain about its original prediction, especially when that prediction is incorrect.}
    \label{fig:quali_high_VPD}
\end{figure}

\begin{figure}
    \centering
    \includegraphics[width=\linewidth]{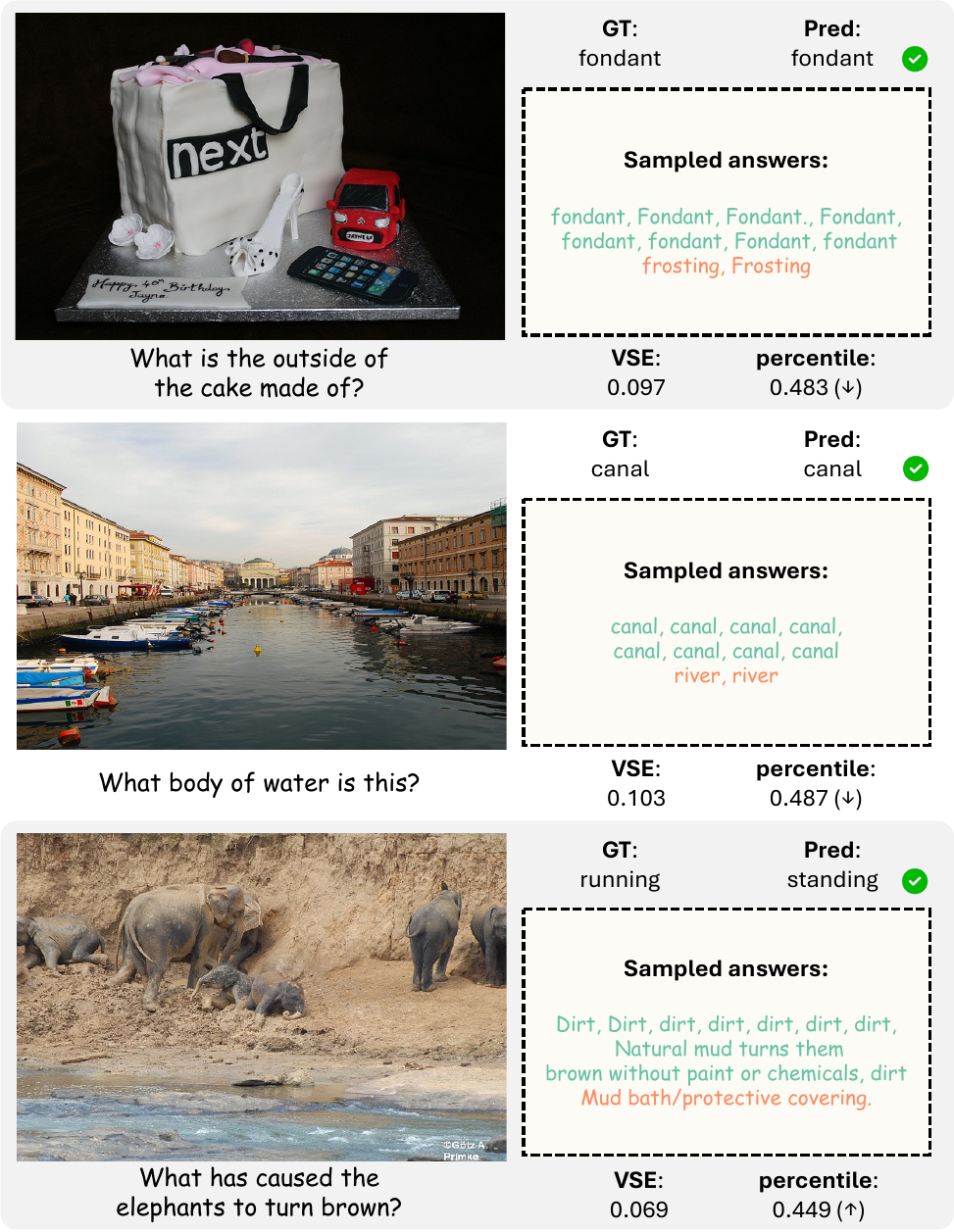}
    \caption{\textbf{Qualitative Results.} A low VSE indicates that the model is certain about its original prediction, especially when that prediction is correct.}
    \label{fig:quali_low_VPD}
\end{figure}

\begin{figure}[t]
    \centering
    \includegraphics[width=\linewidth]{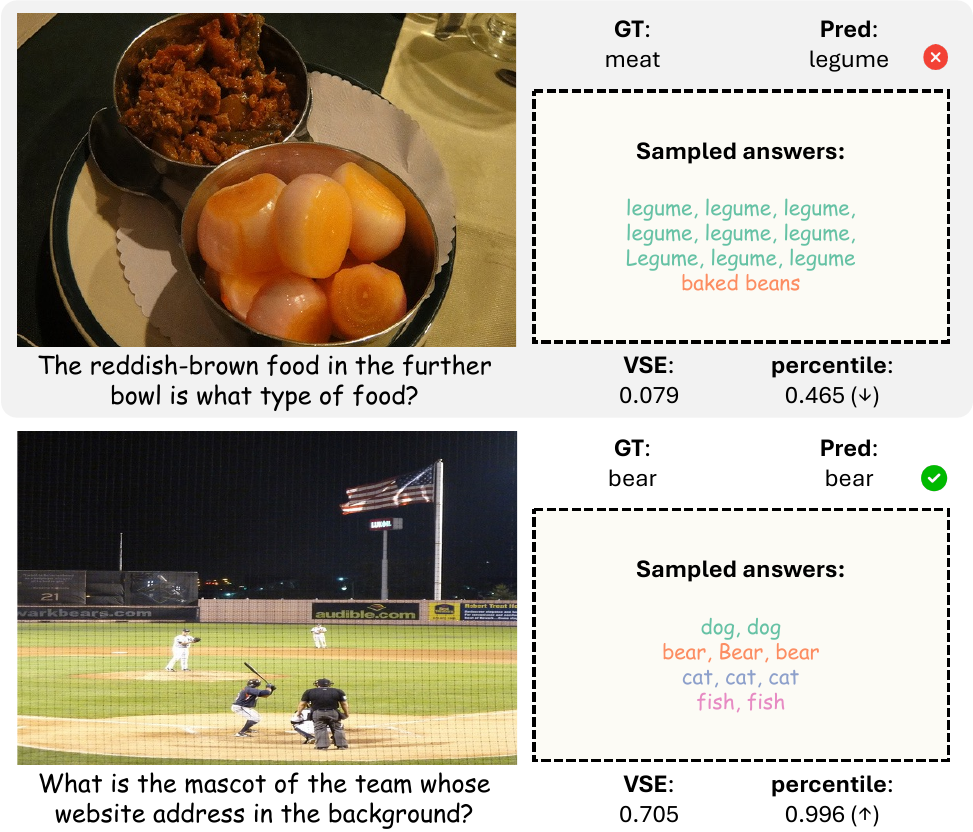}
    \caption{\textbf{Qualitative Results.} Failure cases where VSE is low for incorrect answer (top) and high for correct answer (bottom). In the bottom example, the visual evidence is highly ambiguous, suggesting that the model arrives at the correct answer largely by chance.}
    \label{fig:quali_incorrects}
\end{figure}

We provide qualitative results where high VSE indicates high uncertainty of wrong answers in Fig.~\ref{fig:quali_high_VPD}; low VSE indicates low uncertainty or confident of correct answers in Fig.~\ref{fig:quali_low_VPD}; and failure cases of VSE in Fig.~\ref{fig:quali_incorrects}. Percentile indicates the rank of the VSE values within the dataset to reflect a value within a bounded range $[0,1]$ for ease of interpretability. Results are shown for AOKVQA dataset using Qwen2.5-VL model.

\clearpage
\bibliographystyle{splncs04}
\bibliography{main}

\end{document}